\documentclass{article}
\usepackage{ijcai24}

\usepackage{multirow}
\usepackage{multicol}
\usepackage{times}
\usepackage{amsfonts,amssymb}
\usepackage{soul}
\usepackage{url}
\usepackage[hidelinks]{hyperref}
\usepackage[utf8]{inputenc}
\usepackage[small]{caption}
\usepackage{graphicx}
\usepackage{amsmath}
\usepackage{amsthm}
\usepackage{comment}
\usepackage{booktabs}
\usepackage{algorithm}
\usepackage{algpseudocode}
\usepackage[switch]{lineno}
\usepackage{colortbl}
\usepackage{makecell}
\usepackage[table]{xcolor}
\usepackage{tabularx}
\usepackage{marvosym}
\usepackage{microtype}
\usepackage{subfig}
\usepackage{hyperref}

\title{VCformer: Variable Correlation Transformer with Inherent Lagged Correlation for Multivariate Time Series Forecasting}

\author{
Yingnan Yang$^1$
\and
Qingling Zhu$^2$\And
Jianyong Chen$^{1,}$\thanks{Corresponding author}
\affiliations
 $^1$College of Computer Science and Software Engineering, Shenzhen University, Shenzhen, China\\
 $^2$National Engineering Laboratory for Big Data System Computing Technology, Shenzhen University, Shenzhen, China\\
\emails
csyyn1009@gmail.com,
\{zhuqingling, jychen\}@szu.edu.cn
}

\begin{document}

\maketitle

\begin{abstract}

 Multivariate time series (MTS) forecasting has been extensively applied across diverse domains, such as weather prediction and energy consumption. 
 However, current studies still rely on the vanilla point-wise self-attention mechanism to capture cross-variable dependencies, which is inadequate in extracting the intricate cross-correlation implied between variables. To fill this gap, we propose \textbf{V}ariable \textbf{C}orrelation Transformer (VCformer), which utilizes \textbf{V}ariable \textbf{C}orrelation \textbf{A}ttention (VCA) module to mine the correlations among variables. Specifically, based on the stochastic process theory, VCA calculates and integrates the cross-correlation scores corresponding to different lags between queries and keys, thereby enhancing its ability to uncover multivariate relationships. Additionally, inspired by Koopman dynamics theory, we also develop \textbf{K}oopman \textbf{T}emporal \textbf{D}etector (KTD) to better address the non-stationarity in time series. The two key components enable VCformer to extract both multivariate correlations and temporal dependencies. Our extensive experiments on eight real-world datasets demonstrate the effectiveness of VCformer, achieving top-tier performance compared to other state-of-the-art baseline models. Code is available at this repository: \href{https://github.com/CSyyn/VCformer}{https://github.com/CSyyn/VCformer}.
\end{abstract}

\begin{figure*}[t]
    \centering
    \subfloat[Analogous to $attn(v_1,v_2)$]{
        \includegraphics[width=0.32\textwidth]{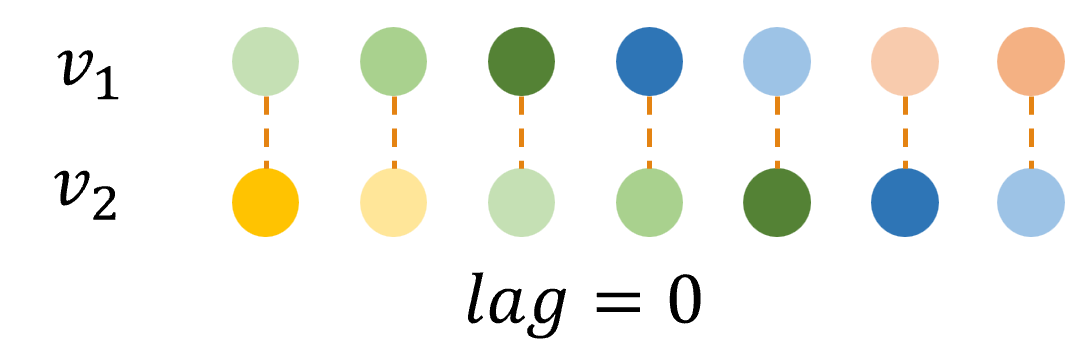}
        \label{fig:dot-product}
        
    }
    \hfill
    \subfloat[Different lags inherent in time series]{
            \includegraphics[width=0.64\textwidth]{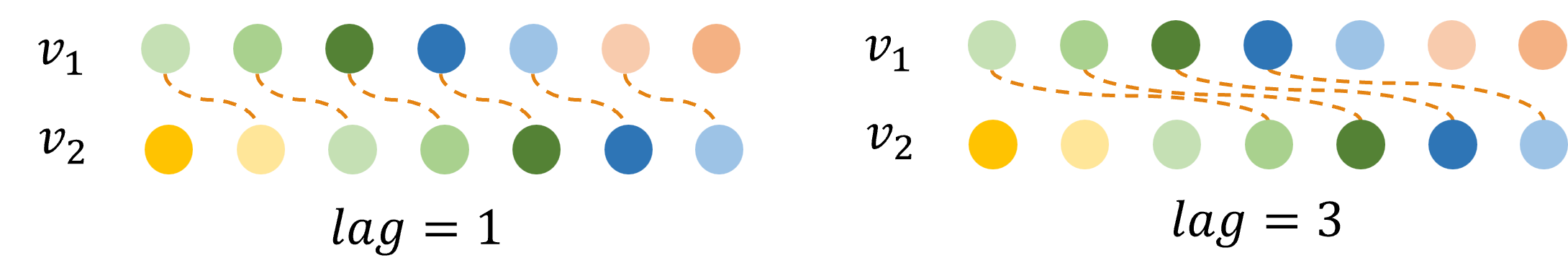}
            \label{fig:lag}
    }
    \caption{Illustration of the dot-product method to obtain correlations with different lags. For example, $lag=3$ shows the similarity between $v_1$ and $v_2$ (points with the same color)} 
    \label{fig:motivation}
\end{figure*}
\section{Introduction}

    Multivariate time series (MTS) forecasting is widely used in a range of applications, including energy consumption, weather, traffic, economics, and other fields \cite{energy,economics,traffic,other_fields,weather}. Unlike univariate time series, MTS involves multiple interrelated time-dependent variables, presenting unique challenges in capturing intricate inter-variable dependencies \cite{Revisiting_CI_and_CD}. Consequently, MTS forecasting has always been a prominent research domain in both industry and academia.

    The achievement of Transformer \cite{Transformer} in natural language processing \cite{NLP} has led to the emergence of numerous Transformer variants for time series prediction tasks. 

    These models have developed various sophisticated attention mechanisms and enhancements to the Transformer architecture \cite{Informer,Autoformer,LogTrans,FEDformer}, which demonstrate a remarkable modelling ability for temporal dependencies in time series data \cite{Trans_in_TS_survey}.
    
    However, there is an ongoing academic discourse regarding their ability to effectively capture temporal dependencies \cite{DLinear} which typically embed each time step into a mix-channel token and apply attention mechanism on every token. Considering that these methods may overlook the valuable multivariate relationships, which is crucial for MTS forecasting, researchers have begun to focus on ensuring the channel independence and incorporating mutual information to explicitly model multivariate correlations. \cite{Crossformer,PatchTST,DSformer,iTransformer}. 
    
    Nevertheless, the traditional self-attention mechanism obtain the relationship between two variables via dot-product which can be approximately analogous to $attn(v_1,v_2)$ shown in Figure \ref{fig:dot-product}. This approach aligns each time step of two variables ignoring the potential existence of different time delays between them, as shown in Figure \ref{fig:lag} \cite{lagged_correaltion_analysis_1,lagged_correaltion_analysis_2}.

    Addressing the limitations of vanilla variable point-wise attention, we introduce the \textbf{V}ariable \textbf{C}orrelation Transformer (VCformer) to fully exploit lagged correlation inherent in MTS through the \textbf{V}ariable \textbf{C}orrelation \textbf{A}ttention (VCA) module. The VCA module calculates the global strength of correlations between each query and key across different feature. Inspired by stochastic process theory \cite{stochastic,time_series_analysis}, it not only computes auto-correlations akin to those in Autoformer \cite{Autoformer} but also extends this concept to determine lagged cross-correlations among various variates. The method employs a $\operatorname{ROLL}$ operation combined with Hadamard products to approximate these lagged correlations effectively. Furthermore, VCA adaptively aggregates lagged correlation over various lag lengths, thereby determining the comprehensive correlation for each variate. To enhance the model's capability in addressing the non-stationary property in MTS, we design the \textbf{K}oopman \textbf{T}emporal \textbf{D}etector (KTD) module inspired by Koopman theory in dynamics \cite{Koopman_1}. Experimentally, VCformer achieves state-of-the-art (SOTA) performance on eight real-world datasets. We also conduct experiments about VCA generality on other previous SOTA Transformer-based models, which demonstrates the powerful capability of modelling channel dependencies of VCA. In general, our contributions lie in the following three aspects:\\
    \begin{itemize}
        \item We propose a novel model for MTS forecasting, which is called VCformer. It learns both variable correlations and temporal dependencies of MTS. 
        \item We design VCA mechanism to fully exploit lagged correlations among different variates. Additionally, we propose KTD inspired by Koopman theory in dynamics to effectively address non-stationarity in MTS forecasting.
        \item Experimentally, VCformer achieves top-tier performance on eight real-world datasets. To further evaluate the generality of VCA function, VCA is used in other Transformer-based models and gets better performance.
    \end{itemize}

\begin{figure*}[t]
    \centering 
    \resizebox{\textwidth}{!}{\includegraphics{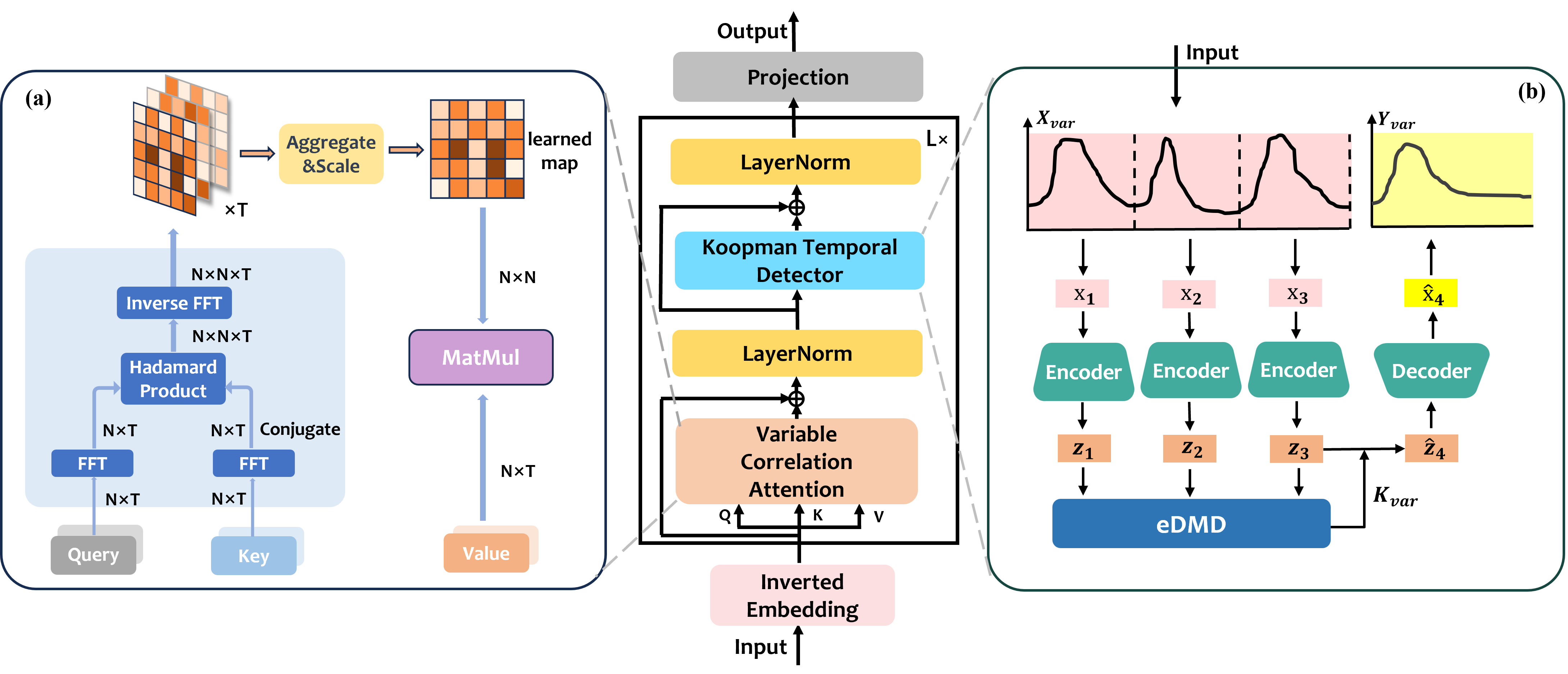}}
    \caption{Overall structure of VCformer, VCA module (a) and KTD module (b)}
    \label{fig:model}
\end{figure*}

\section{Related Work}

Advancing beyond contemporaneous Temporal Convolutional Networks \cite{DeepGLO} and RNN-based models \cite{DeepAR,LSTNet}, Transformer variants have shown excellent capability in sequence modelling. All the modifications can be divided into two groups according to their focus: solely on modeling temporal dependencies and addressing both temporal and variable dependencies.

For the former, a series of sophisticated attention mechanisms has been developed which can be roughly classified into three categories. The first category is to remove redundant information by introducing sparse bias, thereby reducing the quadratic complexity of vanilla Transformer \cite{LogTrans,Informer}. The second is to transfer the self-attention mechanism from time domain to frequency domain. This shift is facilitated by 
Fast Fourier Transform or other frequency analysis tools, enabling a more granular extraction for temporal dependencies at sub-series level \cite{Autoformer,FEDformer}. The third category is related to tackling the distribution shift phenomenon in time series such as De-stationary attention \cite{NSTrans}. Beyond these attention-focused innovations, the former also include methods that incorporate multi-resolution analysis of time series via hierarchical architectures \cite{Pyraformer}.

With the primary focus on extracting temporal dependencies, these models designed various exquisite attention mechanisms and fancy architectures. However, a critical vulnerability in these models is the neglect of the rich cross-variable information, which is important for MTS forecasting tasks.

For the later in addressing MTS forecasting, two primary dimensions emerge in multivariate modeling: Channel-Independent (CI) and Channel-Dependent (CD). CI takes variates of time series independently and adopts the shared backbone. CD predicts future values by taking into account all the channels \cite{revisiting_LSTF}. \cite{PatchTST} introduces patching and CI strategies, significantly enhancing its performance within Transformer-based architectures. Although CI structure is simple, its time-consuming training and inference has catalyzed the development of CD method for modeling multivariate relationships. For the CD method, \cite{Crossformer} employs temporal and variable attention serially to capture both cross-time and cross-dimension dependencies, while \cite{DSformer} applies them in parallel. Moreover, iTransformer \cite{iTransformer} revolutionizes the vanilla Transformer by inverting the duties of the traditional attention mechanism and the feed-forward network. They focus on capturing multivariate correlations and learning nonlinear representations respectively.

While these above works acknowledge the significance of modelling multivariate relationships, they adopt the classical self-attention mechanism based on point-wise method, which does not fully exploit the relationship among variable sequences. Despite the existing methods for analysis of lagged cross-correlations in time series \cite{lagged_correaltion_analysis_1,lagged_correaltion_analysis_2,lagged_correaltion_analysis_3}, these time series Transformers in the literature have not leveraged them among variables, thereby limiting their predictive performance.

\section{Method}
    
    In MTS forecasting, given historical observations $\mathbf{X}=\left\{\mathbf{x}_1, \ldots, \mathbf{x}_T\right\} \in \mathbb{R}^{T \times N}$ with \textit{T} time steps and \textit{N} variates, we predict the future \textit{H} time steps $\mathbf{Y}=\left\{\mathbf{x}_{T+1}, \ldots, \mathbf{x}_{T+H}\right\} \in \mathbb{R}^{H \times N}$. To tackle this MTS forecasting task, we proposes \textbf{VCformer} which is shown in Figure \ref{fig:model}. 
\subsection{Background}
    In this section, we first discuss the current limitation of vanilla variable attention in modelling feature-wise dependencies. This then motivates us to propose the variable cross-correlation attention mechanism, which operates across the feature channels for learning cross-correlation among variates.

    Next, we review the Koopman theory and treat time series as dynamics. Based on this, we design the KTD module and combine it with the variable cross-correlation attention to learn both channels and time-steps dependencies.
\subsubsection{Limitation of Vanilla Variable Attention}    
    In the previous Transformer-based forecasters which adopted attention mechanism for facilitating the temporal dependencies, the self-attention module employs the linear projections to get queries, keys and values $\mathbf{Q}, \mathbf{K}, \mathbf{V} \in \mathbb{R}^{T \times D}$, where ${D}$ is the projected dimension.
    With the queries $Q=\left[\mathbf{q}_1, \mathbf{q}_2, \ldots, \mathbf{q}_T\right]^{\top}$ and keys $K=\left[\mathbf{k}_1, \mathbf{k}_2, \ldots, \mathbf{k}_T\right]^{\top}$, the pre-Softmax attention score is the computation with $\mathbf{A}_{i, j}=\left(\mathbf{Q} \mathbf{K}^{\top} / \sqrt{D}\right)_{i, j} \propto \mathbf{q}_i^{\top} \mathbf{k}_j$. Nevertheless, feature-wise information, where each of the ${D}$ features corresponds to an entry of $\mathbf{q}_i \in \mathbb{R}^{1 \times D}$ or $\mathbf{k}_j \in \mathbb{R}^{1 \times D}$, is absorbed into such inner-product representation. This thus makes such temporal attention unable to explicitly leverage the feature-wise information.  
    iTransformer \cite{iTransformer} considered the limitation of temporal attention and proposed the inverted Transformer to capture cross-variable dependencies that instead computes $K^{\top}Q \in \mathbb{R}^{{D} \times {D}}$. This simple design is suitable for capturing instantaneous cross-correlation, but it is insufficient for MTS data which is coupled with the intrinsic temporal dependencies. In particular, the variates of MTS data can be correlated with each other, yet with a lag interval. This phenomenon is referred to as lagged cross-correlation in MTS analysis \cite{lagged_correaltion_analysis_1,lagged_correaltion_analysis_2,lagged_correaltion_analysis_3}. Additionally, a variate in MTS data can even be correlated with the delayed copy of itself which is termed auto-correlation \cite{Autoformer}. With yet less-efficient modelling capabilities of cross-correlation, we hereby aim to derive a flexible and efficient correlated attention mechanism that can elevate existing Transformer-based models. 
\subsubsection{Non-linear Dynamics Tackled by Koopman Theory}

    Koopman theory \cite{Koopman_1,Koopman_2} shows that a linear dynamical system can be represented as an infinite-dimensional non-linear Koopman operator $\mathcal{K}$, which operates on a space of measurement functions $g$, such that:
    \begin{equation}
    \mathcal{K} \circ g\left(x_t\right)=g\left(\mathbf{F}\left(x_t\right)\right)=g\left(x_{t+1}\right)
    \end{equation}
    Dynamic Mode Decomposition(DMD) \cite{DMD} seeks the best fitted matrix $K$ to approximate infinite-dimensional operator $\mathcal{K}$ by collecting the observed system states (a.k.a \textit{snapshots}). However, it is highly nontrivial to find appropriate measurement functions $g$ as well as the Koopman operator $\mathcal{K}$. Therefore, by the universal approximation theorem \cite{Approximation_DNN} of deep networks, many works employ DNNs to learn measurement functions in a data-driven way \cite{DNN_Koopman_1,DNN_Koopman_2,DNN_Koopman_3,DNN_Koopman_4,DNN_Koopman_5}.

Koopman theory serves as a connection between finite-dimensional nonlinear dynamics and infinite-dimensional linear dynamics, enabling the use of spectral analysis tools for detailed examination. In this paper, we consider time series data $\mathbf{X}=\left\{\mathbf{x}_1, \ldots, \mathbf{x}_T\right\} $ as observations of a series of dynamic system states, where $\mathbf{x}_i \in \mathbb{R}^{N}$ is the system state. Therefore, we design the KTD module which leverage Koopman-based approaches to tackle nonlinear dynamics.

\subsection{Structure Overview}
The proposed VCformer is shown in Figure \ref{fig:model}. Following the same Encoder-only structure as iTransformer \cite{iTransformer}, we adopt the \text{Inverted Embedding : }$\mathbb{R}^{T} \mapsto \mathbb{R}^{D}$, which regards each univariate time series as the embedded token, instead of embedding multiple variates at the same time as the (temporal) token. By stacking $L$ layers with VCA and KTD modules, the cross-variable relationships and temporal dependencies in time series can be captured. The final prediction is obtained by the \text{Projection : }$\mathbb{R}^{D} \mapsto \mathbb{R}^{H}$.

\subsection{Variable Correlation Attention}

    Our VCA is comprised of lagged cross-correlation calculation and scores aggregation. 
\subsubsection{Lagged Cross-correlation Computing}

Recall from stochastic process theory \cite{stochastic} that for any real discrete-time process $\left\{\mathcal{X}_t\right\}$, its auto-correlation $R_{\mathcal{X}, \mathcal{X}}(\tau)$ can be computed as follows:
\begin{equation}R_{\mathcal{X}, \mathcal{X}}(\tau)=\lim _{L \rightarrow \infty} \frac{1}{L} \sum_{\tau=1}^L \mathcal{X}_t \mathcal{X}_{t-{\tau}}\end{equation}
Given the queries $ Q=\left[\mathbf{q}_1, \mathbf{q}_2, \ldots,\mathbf{q}_{N}\right]$ and keys $ K=\left[\mathbf{k}_1, \mathbf{k}_2, \ldots,\mathbf{k}_{N}\right]$ expressed in feature-wise dimension where 
$\mathbf{q}_i, \mathbf{k}_j \in \mathbb{R}^{T \times 1}$, we make an approximation for the auto-correlation of variates $i$:
\begin{equation}
R_{\mathbf{q}_i, \mathbf{k}_i}(\tau)=\sum_{\tau=1}^T\left(\mathbf{q}_i\right)_{t} \cdot\left(\mathbf{k}_i\right)_{t-{\tau}}=\mathbf{q}_i \odot \operatorname{ROLL}\left(\mathbf{k}_i,{\tau} \right)
\label{eq:auto_cor_variate}
\end{equation}
where $\operatorname{ROLL}\left(\mathbf{k}_i,{\tau} \right)$ denotes the elements of $\mathbf{k}_i$ shift along the time dimension and $\odot$ denotes the Hadamard product. This idea was also harnessed in Autoformer \cite{Autoformer}. Similarly, we can compute the cross-correlation between variate $i$ and $j$ by:
\begin{equation}\operatorname{LAGGED-COR}\left(\mathbf{q}_i,\mathbf{k}_j\right)=\mathbf{q}_i \odot \operatorname{ROLL}\left(\mathbf{k}_j,{\tau} \right)
\label{lagged_cor}
\end{equation}
where $\tau \in \left[1,T\right]$. Consequently, we calculate all the variates lagged cross-correlations with different lag lengths in this way.

\subsubsection{Scores Aggregation}

To obtain the total correlation of variate $i$ and $j$, we aggregate different lags $\tau$ from $1$ to $T$ with learnable parameters $\mathbf{\lambda}=\left[\lambda_1,\lambda_2, \ldots, \lambda_{T}\right]$ to more accurately calculate the effect of lagged correlation:
\begin{equation}\operatorname{COR}\left(\mathbf{q}_i,\mathbf{k}_j\right)=\sum_{\tau=1}^T \lambda_iR_{\mathbf{q}_i, \mathbf{k}_j}(\tau)
\label{score_aggregation}
\end{equation}
Finally, the VCA performs softmax on the learned multivariate correlation map $\mathbf{A} \in \mathbb{R}^{N \times N}$ at each row and obtains the output via:
\begin{equation}\operatorname{VCA}\left(\mathbf{Q},\mathbf{K},\mathbf{V}\right)=\operatorname{SOFTMAX}\left(\operatorname{COR}\left(\mathbf{Q},\mathbf{K}\right)\right)\mathbf{V}\end{equation}

\subsection{Koopman Temporal Detector}
We employ the KTD to address the non-stationarity in the input series $\mathbf{X}_{var} \in \mathbb{R}^{N \times D}$ with multivariate correlation information.
Remarkably, it is non-trivial to directly capture the non-stationarity in the entire series $\mathbf{X}_{var}$, but fortunately we identify that the localized time series exhibits weak stationarity, thereby aligning with the Koopman theory for nonlinear dynamics analysis. Consequently, we divide the input $\mathbf{X}_{var}$ into $\frac{D}{S}$ segments $\mathbf{x}_j$ of length $S$:
\begin{equation}
\mathbf{x}_j = \mathbf{X}_{\text{var}}[:, (j-1)S : jS], \quad j=1, 2, \ldots, \frac{D}{S}.
\end{equation}
where each segment can be served as a snapshot for the system. Subsequently, for every $\mathbf{x}_j \in \mathbb{R}^{N \times S}$, we leverage MLP-based \text{Encoder : } $\mathbb{R}^{N \times S} \mapsto \mathbb{R}^M $ to project it into a Koopman space embedding $\mathbf{z}_j \in \mathbb{R}^{M}$. 
According to eDMD \cite{eDMD}, these embeddings $Z=\left[z_1,z_2, \ldots, z_{\frac{D}{S}}\right] \in \mathbb{R}^{\frac{D}{S} \times M}$ are then utilized to calculate the fitted matrix ${K_{\text{var}}}$, facilitating an approximation of the infinite Koopman operator $\mathcal{K}$.

Specifically, given the Koopman embedding ${Z}$, we construct two matrices $Z_{\text {back }}=\left[z_1, z_2, \ldots, z_{\frac{D}{S}-1}\right] \in \mathbb{R}^{\left(\frac{D}{S}-1\right) \times M}$ and $Z_{\text {fore }}=\left[z_2, z_3, \ldots, z_{\frac{D}{S}}\right] \in \mathbb{R}^{\left(\frac{D}{S}-1\right) \times M}$, which respectively contain information of adjacent embeddings. After that, the fitted matrix $K_{\text{var}} \in \mathbb{R}^{D \times D}$ can be calculated as the following equation:
\begin{equation}
K_{\text {var }}=Z_{\text {fore }} Z_{\text {back }}^{\dagger}
\end{equation}
where $Z_{\text{back }}^{\dagger}$ is the Moore-Penrose inverse of $Z_{\text{back }}$. Following the deviation of $K_{\text{var}}$, we iteratively apply it to predict $\frac{H}{S}$ Koopman embeddings as follows:
\begin{equation}
\hat{z}_{\frac{T}{S}+t}=\left(K_{\text {var }}\right)^t z_{\frac{T}{S}}, \quad t=1,2, \ldots, H / S.
\end{equation}
In this way, a prediction of length $H$ is obtained.
Finally, to obtain the output of KTD, we adopt a \text{Decoder : }$\mathbb{R}^M \mapsto \mathbb{R}^{N \times S}$, which maps the predicted embeddings back, yielding $Y_{\text{var}}$ as follows: 

\begin{equation}
Y_{\text{var}}=\left[\hat{\mathrm{x}}_{\frac{T}{S}+1}, \ldots, \hat{\mathrm{x}}_{\frac{T}{S}+\frac{H}{S}}\right]^{\top}
\end{equation}

\begin{table*}[t]
    \centering
    \resizebox{\textwidth}{!}{
        \begin{tabular}{lccccccccccccccccccc}
            \toprule
            \multicolumn{2}{c}{Method} & \multicolumn{2}{c}{\textbf{VCformer}} & \multicolumn{2}{c}{iTransformer} & \multicolumn{2}{c}{PatchTST} & \multicolumn{2}{c}{DSformer} & \multicolumn{2}{c}{Koopa} & \multicolumn{2}{c}{Crossformer} & \multicolumn{2}{c}{TimesNet} & \multicolumn{2}{c}{DLinear} & \multicolumn{2}{c}{Stationary} \\
            \midrule
            \multicolumn{2}{c}{Metric} & MSE & MAE & MSE & MAE & MSE & MAE& MSE & MAE& MSE & MAE& MSE & MAE & MSE & MAE & MSE & MAE & MSE & MAE \\
            
            \midrule
            \multirow{4}{*}{\rotatebox{90}{Weather}} 
            & 96  &  \textcolor{blue}{\underline{0.171}} & \textcolor{blue}{\underline{0.220}} & 0.174 & 0.214 & 0.177 & 0.218& \textcolor{red}{\textbf{0.170}} & \textcolor{red}{\textbf{0.217}} & 0.177 & 0.226& 0.185 & 0.248 &  0.172 & 0.220 & 0.196 & 0.255 & 0.205 & 0.265 \\
            & 192 &  0.230 & 0.266 & \textcolor{red}{\textbf{0.221}} & \textcolor{red}{\textbf{0.254}} & 0.242 & 0.271 & 0.253& 0.296 & \textcolor{blue}{\underline{0.223}}& \textcolor{blue}{\underline{0.257}} & 0.229& 0.305 & 0.230 &  0.281  & 0.253 & 0.323 & 0.233 & 0.274 \\
            & 336 &  \textcolor{blue}{\underline{0.280}} & \textcolor{blue}{\underline{0.299}} & \textcolor{red}{\textbf{0.278}} & \textcolor{red}{\textbf{0.296}} & 0.290 & 0.305& 0.285 & 0.310& 0.281 & 0.299 & 0.323 & 0.285 &  0.335 & 0.311 & 0.325 & 0.390 & 0.296 & 0.317 \\
            & 720 &  \textcolor{red}{\textbf{0.352}} & \textcolor{red}{\textbf{0.344}} & \textcolor{blue}{\underline{0.354}} & \textcolor{blue}{0.349} & 0.362 & 0.350& 0.395 & 0.391& 0.360 & 0.350 & 0.665 & 0.356 &  0.398 & 0.356 & 0.389 & 0.437 & 0.372 & 0.365 \\
            \midrule
            
            \multirow{4}{*}{\rotatebox{90}{Electricity}} 
            & 96  &  \textcolor{red}{\textbf{0.150}} & \textcolor{red}{\textbf{0.242}} & \textcolor{blue}{\underline{0.154}} & \textcolor{blue}{\underline{0.245}} & 0.188 & 0.280& 0.164 & 0.261 & 0.174 & 0.273& 0.153 & 0.250 &  0.182 & 0.285 & 0.195 & 0.276 & 0.172 & 0.275 \\
            & 192 &  \textcolor{red}{\textbf{0.167}} & \textcolor{red}{\textbf{0.255}} & \textcolor{blue}{\underline{0.169}} & \textcolor{blue}{\underline{0.258}} & 0.193 & 0.285& 0.177 & 0.272& 0.195 & 0.291& 0.223 & 0.329 &  0.247 & 0.329 & 0.194 & 0.280 & 0.187 & 0.287 \\
            & 336 &  \textcolor{red}{\textbf{0.182}} & \textcolor{red}{\textbf{0.270}} & \textcolor{blue}{\underline{0.185}} & \textcolor{blue}{\underline{0.275}} & 0.211 & 0.302& 0.201 & 0.294& 0.216 & 0.310& 0.191 & 0.291 &  0.256 & 0.338 & 0.207 & 0.295 & 0.208 & 0.307 \\
            & 720 &  \textcolor{red}{\textbf{0.221}} & \textcolor{red}{\textbf{0.302}} & \textcolor{blue}{\underline{0.225}} & \textcolor{blue}{\underline{0.308}} & 0.253 & 0.335& 0.242 & 0.327& 0.265 & 0.346& 0.609 & 0.568 &  0.311 & 0.382 & 0.242 & 0.328 & 0.235 & 0.329 \\
            \midrule
           
            \multirow{4}{*}{\rotatebox{90}{Traffic}} 
            & 96  &  \textcolor{red}{\textbf{0.454}} & \textcolor{red}{\textbf{0.310}} & 0.717 & 0.466 & \textcolor{blue}{\underline{0.475}} & \textcolor{blue}{\underline{0.303}}& 0.546 & 0.352& 0.539 & 0.368& 0.530 & 0.285 &  0.592 & 0.315 & 0.640 & 0.388 & 0.732 & 0.418 \\
            & 192 &  \textcolor{red}{\textbf{0.468}} & \textcolor{red}{\textbf{0.315}} & \textcolor{blue}{\underline{0.472}} & \textcolor{blue}{\underline{0.320}} & 0.474 & 0.322& 0.547 & 0.347& 0.552 & 0.375& 0.607 & 0.311 &  0.645 & 0.336 & 0.593 & 0.362 & 0.756 & 0.425 \\
            & 336 &  \textcolor{red}{\textbf{0.486}} & \textcolor{red}{\textbf{0.325}} & \textcolor{blue}{\underline{0.488}} & \textcolor{blue}{\underline{0.330}} & 0.489 & 0.332& 0.562 & 0.352& 0.573 & 0.383& 0.642 & 0.324 &  0.659 & 0.347 & 0.600 & 0.365 & 1.172 & 0.680 \\
            & 720 &  \textcolor{red}{\textbf{0.524}} & \textcolor{red}{\textbf{0.348}} & 0.530 & 0.361 & \textcolor{blue}{\underline{0.526}} & \textcolor{blue}{\underline{0.356}}& 0.597 & 0.370& 0.632 & 0.407& 0.592 & 0.380  & 0.723 & 0.388 & 0.634 & 0.388 & 0.896 & 0.516 \\
            \midrule

            \multirow{4}{*}{\rotatebox{90}{ETTh1}} 
            & 96  &  \textcolor{blue}{\underline{0.376}} & \textcolor{blue}{\underline{0.397}} & 0.380 & 0.398 & 0.378 & \textcolor{red}{\textbf{0.396}}& \textcolor{red}{\textbf{0.373}} & \textcolor{blue}{\underline{0.397}}& 0.389 & 0.403& 0.384 & 0.409 &  0.438 & 0.447 & 0.479 & 0.471 & 0.761 & 0.612 \\
            & 192 &  \textcolor{blue}{\underline{0.431}} & \textcolor{blue}{\underline{0.427}} & 0.433 & 0.428 & 0.433 & 0.427& \textcolor{red}{\textbf{0.419}} & \textcolor{red}{\textbf{0.425}}& 0.438 & 0.431& 0.461 & 0.459 &  0.488 & 0.472 & 0.448 & 0.443 & 0.746 & 0.599 \\
            & 336 &  0.473 & 0.449 & 0.475 & 0.451 & \textcolor{blue}{\underline{0.471}} & \textcolor{blue}{\underline{0.448}}& \textcolor{red}{\textbf{0.457}} & \textcolor{red}{\textbf{0.446}}& 0.479 & 0.451& 0.521 & 0.496 &  0.510 & 0.480 & 0.489 & 0.467 & 0.739 & 0.572 \\
            & 720 &  \textcolor{blue}{\underline{0.476}} & \textcolor{blue}{\underline{0.474}} & 0.486 & 0.480 & \textcolor{red}{\textbf{0.472}} & \textcolor{red}{\textbf{0.471}}& 0.499 & 0.497& 0.486 & \textcolor{blue}{\underline{0.474}}& 0.627 & 0.586 &  0.511 & 0.497 & 0.511 & 0.509 & 0.757 & 0.612 \\
            \midrule    
            
            \multirow{4}{*}{\rotatebox{90}{ETTh2}} 
            & 96  &  \textcolor{red}{\textbf{0.292}} & \textcolor{red}{\textbf{0.344}} & \textcolor{blue}{\underline{0.292}} & \textcolor{blue}{\underline{0.344}} & 0.292 & 0.345& 0.296 & 0.351& 0.306 & 0.355& 0.596 & 0.532 &  0.337 & 0.375 & 0.299 & 0.351 & 0.477 & 0.462 \\
            & 192 &  \textcolor{blue}{\underline{0.377}} & \textcolor{red}{\textbf{0.396}} & \textcolor{red}{\textbf{0.375}} & \textcolor{red}{\textbf{0.396}} & 0.388 & 0.405& 0.399 & 0.414& 0.388 & 0.408& 0.880 & 0.663 &  0.442 & 0.435 & 0.385 & 0.413 & 0.571 & 0.507 \\
            & 336 &  \textcolor{red}{\textbf{0.417}} & \textcolor{red}{\textbf{0.430}} & \textcolor{blue}{\underline{0.418}} & \textcolor{blue}{\underline{0.430}} & 0.427 & 0.436& 0.434 & 0.443& 0.430 & 0.443& 1.988 & 1.097 &  0.476 & 0.468 & 0.511 & 0.490 & 0.608 & 0.534 \\
            & 720 &  \textcolor{red}{\textbf{0.423}} & \textcolor{red}{\textbf{0.443}} & \textcolor{blue}{\underline{0.424}} & \textcolor{red}{\textbf{0.443}} & 0.447 & 0.458 & 0.454 & 0.463& 0.472 & 0.470& 2.526 & 1.285 &  0.496 & 0.484 & 0.741 & 0.603 & 0.508 & 0.487 \\
            \midrule
    
            \multirow{4}{*}{\rotatebox{90}{ETTm1}} 
            & 96  &  \textcolor{red}{\textbf{0.319}} & \textcolor{red}{\textbf{0.359}} & 0.345 & 0.369 & 0.326 & 0.361& 0.326 & 0.364& 0.334 & 0.372& 0.352 & 0.388 &  0.334 & 0.375 & 0.336 & 0.362 & 0.386 & 0.398 \\
            & 192 &  \textcolor{blue}{\underline{0.364}} & \textcolor{blue}{\underline{0.382}} & 0.386 & 0.391 & 0.372 & \textcolor{red}{\textbf{0.381}} & \textcolor{red}{\textbf{0.360}} & \textcolor{blue}{\underline{0.382}}& 0.374 & 0.391& 0.409 & 0.436 &  0.385 & 0.401 & 0.378 & 0.389 & 0.459 & 0.434 \\
            & 336 &  \textcolor{blue}{\underline{0.399}} & \textcolor{blue}{\underline{0.405}} & 0.423 & 0.416 & 0.404 & \textcolor{red}{\textbf{0.403}}& \textcolor{red}{\textbf{0.394}} & \textcolor{blue}{\underline{0.405}}& 0.409 & 0.414 & 0.424 & 0.428 &  0.410 & 0.411 & 0.413 & 0.416 & 0.551 & 0.485 \\
            & 720 &  \textcolor{red}{\textbf{0.467}} & \textcolor{blue}{\underline{0.442}} & 0.491 & 0.445 & \textcolor{red}{\textbf{0.467}} & \textcolor{red}{\textbf{0.438}}& 0.474 & 0.451& \textcolor{blue}{\underline{0.473}} & 0.448& 0.569 & 0.528 &  0.513 & 0.473 & 0.475 & 0.454 & 0.585 & 0.516 \\
            \midrule
            
            \multirow{4}{*}{\rotatebox{90}{ETTm2}} 
            & 96  &  \textcolor{red}{\textbf{0.180}} & \textcolor{red}{\textbf{0.266}} & 0.190 & 0.276 & 0.193 & 0.280& 0.201 & 0.286& 0.187 & 0.271& 0.297 & 0.370 &  \textcolor{blue}{\underline{0.185}} & \textcolor{blue}{\underline{0.267}} & 0.188 & 0.284 & 0.240 & 0.320 \\
            & 192 &  \textcolor{red}{\textbf{0.245}} & \textcolor{red}{\textbf{0.306}} & 0.251 & 0.311 & \textcolor{blue}{\underline{0.246}} & \textcolor{blue}{\underline{0.307}}& 0.281 & 0.335& 0.253 & 0.314& 0.499 & 0.492 &  0.249 & \textcolor{red}{\textbf{0.306}} & 0.259 & 0.337 & 0.314 & 0.367 \\
            & 336 &  \textcolor{red}{\textbf{0.307}} & \textcolor{red}{\textbf{0.345}} & 0.315 & 0.352 & \textcolor{blue}{\underline{0.314}} & 0.351& 0.336 & 0.367& 0.323 & 0.358& 0.597 & 0.684 &  \textcolor{blue}{\underline{0.314}} & \textcolor{blue}{\underline{0.346}} & 0.334 & 0.389 & 0.340 & 0.371 \\
            & 720 &  \textcolor{red}{\textbf{0.406}} & \textcolor{blue}{\underline{0.402}} & 0.413 & 0.404 & \textcolor{blue}{\underline{0.410}} & 0.405& 0.430 & 0.417& 0.416 & 0.407& 0.835 & 0.659 &  0.411 & \textcolor{red}{\textbf{0.399}} & 0.463 & 0.466 & 0.438 & 0.421 \\
            \midrule
            
            \multirow{4}{*}{\rotatebox{90}{Exchange}} 
            & 96  &  \textcolor{red}{\textbf{0.085}} & \textcolor{red}{\textbf{0.205}} & \textcolor{blue}{\underline{0.090}} & \textcolor{blue}{\underline{0.211}} & 0.100 & 0.231& 0.092 & 0.216& 0.092 & 0.217& 0.139 & 0.265 &  0.108 & 0.244 & 0.110 & 0.266 & 0.154 & 0.297 \\
            & 192 &  \textcolor{red}{\textbf{0.176}} & \textcolor{red}{\textbf{0.299}} & 0.186 & 0.307 & 0.215 & 0.344& 0.189 & 0.312& \textcolor{blue}{\underline{0.182}} & \textcolor{blue}{\underline{0.304}}& 0.241 & 0.375 &  0.278 & 0.391 & 0.218 & 0.376 & 0.374 & 0.447 \\
            & 336 &  \textcolor{red}{\textbf{0.328}} & \textcolor{red}{\textbf{0.415}} & \textcolor{blue}{\underline{0.339}} & \textcolor{blue}{\underline{0.424}} & 0.403 & 0.473& 0.348 & 0.430& 0.349 & 0.432& 0.392 & 0.468 &  0.523 & 0.556 & 0.387 & 0.497 & 0.548 & 0.563 \\
            & 720 &  \textcolor{red}{\textbf{0.830}} & \textcolor{red}{\textbf{0.688}} & 0.898 & 0.718 & 1.057 & 0.782& 0.947 & 0.740& 1.178 & 0.830& 1.11 & 0.802 &  1.224 & 0.856 & \textcolor{blue}{\underline{0.839}} & \textcolor{blue}{\underline{0.695}} & 0.987 & 0.777 \\
            \bottomrule
        \end{tabular}
    }
    \caption{Multivariate long-term time series forecasting results}
    \label{tab:experiment}
\end{table*}

\subsection{Efficient Computation}
For each vector pair $\mathbf{q}_i, \mathbf{k}_j \in \mathbb{R}^{T \times 1}$, the time complexity of the lag-correlation (Equation \ref{score_aggregation}) is $\mathcal{O}\left(T^2\right)$. Therefore, calculating $\operatorname{COR}\left(q_i,\mathbf{K}\right)$ demands $\mathcal{O}\left(NT^2\right)$ time. It leads the overall complexity of VCA to $\mathcal{O}\left(N^2T^2\right)$ in its current form. To alleviate the computational burden, we apply Fast Fourier Transforms (FFT) based on Wiener-Khinchin theorem \cite{wiener}, thus reducing the complexity to $\mathcal{O}\left(N^2T\log T\right)$. Specifically, for computing the lag-correlation in Equation \ref{lagged_cor}, given discrete time series $\left\{\mathcal{X}_t\right\}$ and $\left\{\mathcal{Y}_t\right\}$, the $R_{\mathcal{X}\mathcal{Y}}(\tau)$ can be calculated via FFT as follows:
\begin{equation}
\begin{aligned}
\mathcal{S}_{\mathcal{X Y}}(f) & =\mathcal{F}\left(\mathcal{X}_t\right) \mathcal{F}^*\left(\mathcal{Y}_t\right) \\
&=\int_{-\infty}^{+\infty} \mathcal{X}_t e^{-i 2 \pi t f} d t \overline{\int_{-\infty}^{+\infty} \mathcal{Y}_t e^{-i 2 \pi t f} d t} \\
R_{\mathcal{X} \mathcal{Y}}(\tau)& =\mathcal{F}^{-1}\left(\mathcal{S}_{\mathcal{X Y}}(f)\right)=\int_{-\infty}^{+\infty} \mathcal{S}_{\mathcal{X Y}}(f) e^{i 2 \pi f l} d f,
\end{aligned}
\label{Winner theorm}
\end{equation}
where $\tau \in \left[1,T\right]$. $\mathcal{F}$ and $\mathcal{F}^{-1}$ denotes the FFT and its inverse respectively, and $*$ is the conjugate operation. Specifically, we transform $\mathbf{Q}$ and $\mathbf{K}$ into frequency domain using FFT. Then element-wise multiplication ($a.k.a$ Hadamard Product) is applied to $i$th row of the $\mathcal{F}\left(\mathbf{Q}\right)$ and $\mathcal{F}\left(\mathbf{K}\right)$ to compute the $\operatorname{LAGGED-COR}\left(\mathbf{q}_i,\mathbf{K}\right)$. Extending this process to the entire matrix $\mathcal{F}\left(\mathbf{Q}\right)$ and applying inverse FFTs to these products yield the complete lag-correlations between $\mathbf{Q}$ and $\mathbf{K}$. As FFT and inverse FFT each requires 
$\mathcal{O}\left(T \log T\right)$, the optimized VCA achieves a complexity of  $\mathcal{O}\left(N^2 T \log T\right)$.

\begin{table*}[t]
    \centering
    \resizebox{\textwidth}{!}{
        \begin{tabular}{@{}c|c|cccccccccccc@{}}
                    \toprule
                    \multicolumn{2}{c}{Models}& \multicolumn{2}{c}{\makecell{Nonstationary\\ (2022)}} & \multicolumn{2}{c}{\makecell{Nonstationary\\ (VCA)}} & \multicolumn{2}{c}{\makecell{Autoformer\\(2022)}} & \multicolumn{2}{c}{\makecell{Autoformer\\ (VCA)}} & \multicolumn{2}{c}{\makecell{Informer\\(2021)}} & \multicolumn{2}{c}
                    {\makecell{Informer\\(VCA)}}\\
                    \midrule
                    \multicolumn{2}{c}{Metrics}& MSE & MAE & MSE & MAE & MSE & MAE & MSE & MAE & MSE & MAE  & MSE & MAE\\
                    \midrule
                    \multirow{4}{*}{Electricity} 
                    & 96   & 0.172 & 0.275 & 0.163 & 0.262 & 0.201 & 0.317 & 0.170 & 0.273 & 0.274 & 0.368 & 0.195 & 0.301 \\
                    & 192  & 0.187 & 0.287 & 0.175 & 0.270 & 0.222 & 0.334 & 0.195 & 0.290 & 0.296 & 0.386 & 0.210 & 0.315 \\
                    & 336  & 0.208 & 0.307 & 0.195 & 0.286 & 0.231 & 0.338 & 0.200 & 0.295 & 0.300 & 0.394 & 0.231 & 0.339\\
                    & 720  & 0.235 & 0.329 & 0.230 & 0.310 & 0.254 & 0.361 & 0.237 & 0.331 & 0.373 & 0.439 & 0.266 & 0.361\\
                    \midrule
                    & Average & 0.201 & 0.300 & \textbf{0.191(5.1\%)} & \textbf{0.282(6.0\%)} & 0.227 & 0.338 & \textbf{0.201(11.7\%)} & \textbf{0.297(12.1\%)} & 0.311 & 0.397 & \textbf{0.226(27.5\%)} & \textbf{0.329(17.1\%)}\\
                    \midrule
                    \multirow{4}{*}{Exchange} 
                    & 96   & 0.154 & 0.297 & 0.100 & 0.235 & 0.197 & 0.323 & 0.124 & 0.278 & 0.847 & 0.752 & 0.301 & 0.414 \\
                    & 192  & 0.374 & 0.447 & 0.220 & 0.301 & 0.300 & 0.369 & 0.255 & 0.323 & 1.204 & 0.895 & 0.441 & 0.615 \\
                    & 336  & 0.548 & 0.563 & 0.405 & 0.479 & 0.509 & 0.524 & 0.443 & 0.501 & 1.672 & 1.036 & 0.573 & 0.729 \\
                    & 720  & 0.987 & 0.777 & 0.860 & 0.844 & 1.447 & 0.941 & 1.051 & 0.893 & 2.478 & 1.310 & 1.109 & 0.883 \\
                    \midrule
                    & Average & 0.516 & 0.545 & \textbf{0.396(23.2\%)} & \textbf{0.465(14.7\%)} & 0.613 & 0.517 & \textbf{0.468(23.6\%)} & \textbf{0.499(3.5\%)} & 1.550 & 0.998 & \textbf{0.606(60.9\%)} & \textbf{0.660(33.9\%)}\\
                    \midrule
                    
                    \multirow{4}{*}{Traffic} 
                    & 96   & 0.612 & 0.338 & 0.540 & 0.321 & 0.613 & 0.388 & 0.559 & 0.357 & 0.719 & 0.391 & 0.590 & 0.371\\
                    & 192  & 0.613 & 0.340 & 0.548 & 0.324 & 0.616 & 0.379 & 0.563 & 0.355 & 0.696 & 0.379 & 0.601 & 0.381\\
                    & 336  & 0.618 & 0.328 & 0.554 & 0.331 & 0.622 & 0.337 & 0.570 & 0.366 & 0.777 & 0.420 & 0.595 & 0.382\\
                    & 720  & 0.653 & 0.355 & 0.579 & 0.362 & 0.660 & 0.408 & 0.601 & 0.385 & 0.864 & 0.472 & 0.622 & 0.407\\
                    \midrule
                    & Average & 0.624 & 0.420 & \textbf{0.555(11.6\%)} & \textbf{0.334(20.4\%)} & 0.628 & 0.379 & \textbf{0.573(8.8\%)} & \textbf{0.365(3.5\%)} & 0.764 & 0.416 & \textbf{0.602(21.2\%)} & \textbf{0.385(7.4\%)}\\
                    \bottomrule
        \end{tabular}
    }
    \caption{VCA Generality and improvement for other Transformer-based models}
    \label{tab:generality}
\end{table*}

\begin{table*}[t]
    \centering
    \resizebox{0.75\linewidth}{!}{
    \renewcommand\arraystretch{1.4}
        \begin{tabular}{c|cc|cc|cc|cc|cc} 
            \toprule
             Design & \multicolumn{2}{c}{VCformer} &\multicolumn{2}{c}{Replace VCA} & 
             \multicolumn{2}{c}{w/o VCA} & \multicolumn{2}{c}{Replace KTD} & \multicolumn{2}{c}{w/o KTD} \\
             \midrule
             Metric & MSE & MAE & MSE & MAE & MSE & MAE & MSE & MAE & MSE & MAE \\
             \midrule
             Exchange & \textbf{0.360} & \textbf{0.402} & 0.390 & 0.445 & 0.419 & 0.454 & 0.425 & 0.480 & 0.440 & 0.506 \\
             Traffic & \textbf{0.483} & \textbf{0.324} & 0.527 & 0.351 & 0.498 & 0.365 & 0.498 & 0.337 & 0.518 & 0.351 \\
             Electricity & \textbf{0.180} & \textbf{0.198} & 0.290 & 0.445 & 0.184 & 0.288 & 0.184 & 0.288 & 0.190 & 0.280 \\
             Weather & \textbf{0.258} & \textbf{0.282} & 0.265 & 0.285 & 0.264 & 0.290 & 0.264 & 0.290 & 0.269 & 0.291 \\
             \bottomrule
        \end{tabular}
    }
   \caption{Ablations on VCformer. We conduct substitution and removal experiments on two key components (VCA \& KTD) of VCformer respectively. For the substitution experiments, we replace the VCA and KTD modules with self-attention and FFN module respectively. The average results with all prediction lengths are presented in here.}
   \label{tab:ablation_study}
\end{table*}

\section{Experiment}
\paragraph{Dataset} 
 We conduct extensive experiments on eight widely-used real-world datasets \cite{Informer}, including Electricity Transformer Temperature (ETT) with its four sub-datasets (ETTh1, ETTh2, ETTm1, ETTm2), Weather, Electricity, Traffic and Exchange. Following \cite{Informer}, we adopt the same train/val/test sets with splits ratio 6:2:2. For the ETT datasets, we split the remaining four sub-datasets by the ratio of 7:1:2 following \cite{Autoformer}.
\paragraph{Baselines}
We carefully select a range of SOTA methods as baselines to provide a comprehensive comparison with our proposed approach including (1) Transformer-based methods: Stationary \cite{NSTrans}, Crossformer \cite{Crossformer}, DSformer \cite{Crossformer}, iTransformer \cite{iTransformer}; (2) MLP-based methods: DLinear \cite{DLinear}, Koopa \cite{Koopa}; (3) TCN-based methods: TimesNet \cite{TimesNet}. 
\paragraph{Setups}

 Following \cite{Informer}, we normalize the train/val/test to zero-mean using the mean and standard deviation from the training set. The Mean Square Error (MSE) and Mean Absolute Error (MAE) are selected as evaluation metrics, consistent with previous works. All of models adopt the same prediction length $H=\left\{96,192,336,720\right\}$. For the look-back window with length $T$, we follow the same setting as TimesNet \cite{TimesNet} which sets $T=96$ for all the baselines to ensure the fairness.

\subsection{Time Series Forecasting}

Table \ref{tab:experiment} shows the comprehensive experimental results, where the lower MSE/MAE indicates the more accurate result. And we highlight the best in red and bold, while the second in blue and underlined. As we can see, Table \ref{tab:experiment} illustrate that VCformer consistently achieves top-tier performance across a range of datasets, outperforming other previous SOTA models. It can be attributed to the robust capability of VCA component in extracting correlations among multiple variables. Additionally, it is noteworthy that VCformer achieves the best results on the Exchange dataset which is characterized by high non-stationarity. This success can be owing to the KTD component which augments the power in capturing the non-stationarity from time series. Furthermore, in other datasets like ETT where VCformer does not attain the best benchmark, it still yield competitive results. We also find that the conventional Transformer-based models such as Non-stationary Transformer \cite{NSTrans}, only achieve the modest performance. It further substantiates the previously discussed limitations of the vanilla attention mechanism in tackling temporal dependencies. 

\subsection{VCA Generality}
To further explore the effectiveness of VCA, we migrate the VCA module to several well-known Transformer variants: Non-Stationary Transformer \cite{NSTrans}, Autoformer \cite{Autoformer} and Informer \cite{Informer}. 
Since these Transformer-based models have masked decoders in which the partial values of scores ($\mathbf{Q}\mathbf{K^T}$) are replaced with $-\infty$, the FFT in VCA module can not be used to quickly calculate the lag-correlations of queries and keys. 
 
Therefore, we retain the original design of decoder and simply replace self-attention in encoder with VCA. 
The experimental results are shown in Table \ref{tab:generality}, where (VCA) represents the replaced model. We can see that the VCA module has significantly improved the performance of traditional Transformer-based models (Non-stationary Transformer, Autoformer and Informer), which yields an overall relative MSE reduction with \textbf{13.3\%}, \textbf{14.7\%} and \textbf{36.5\%}. 

\begin{figure}
    \centering
    \resizebox{0.49\textwidth}{!}{\includegraphics{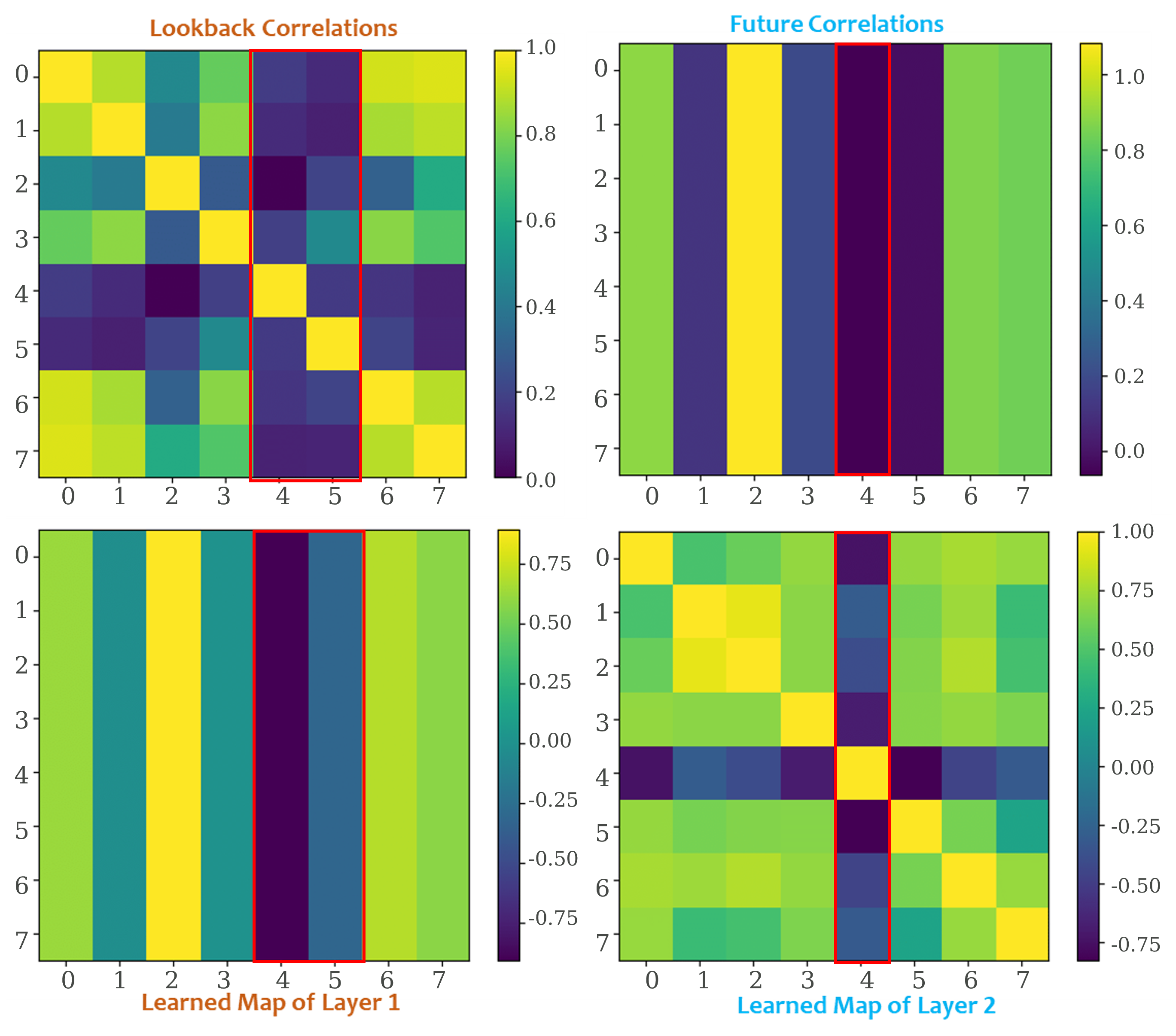}}
    \caption{A case visualization for multivariate correlation analysis. The upper part is the multivariate correlation of past series and future series. The bottom part is the learned correlation maps in different layers.}
    \label{fig:learned_map}
\end{figure}

\begin{figure*}[t]
\centering
\subfloat{
    \includegraphics[width=0.49\textwidth]{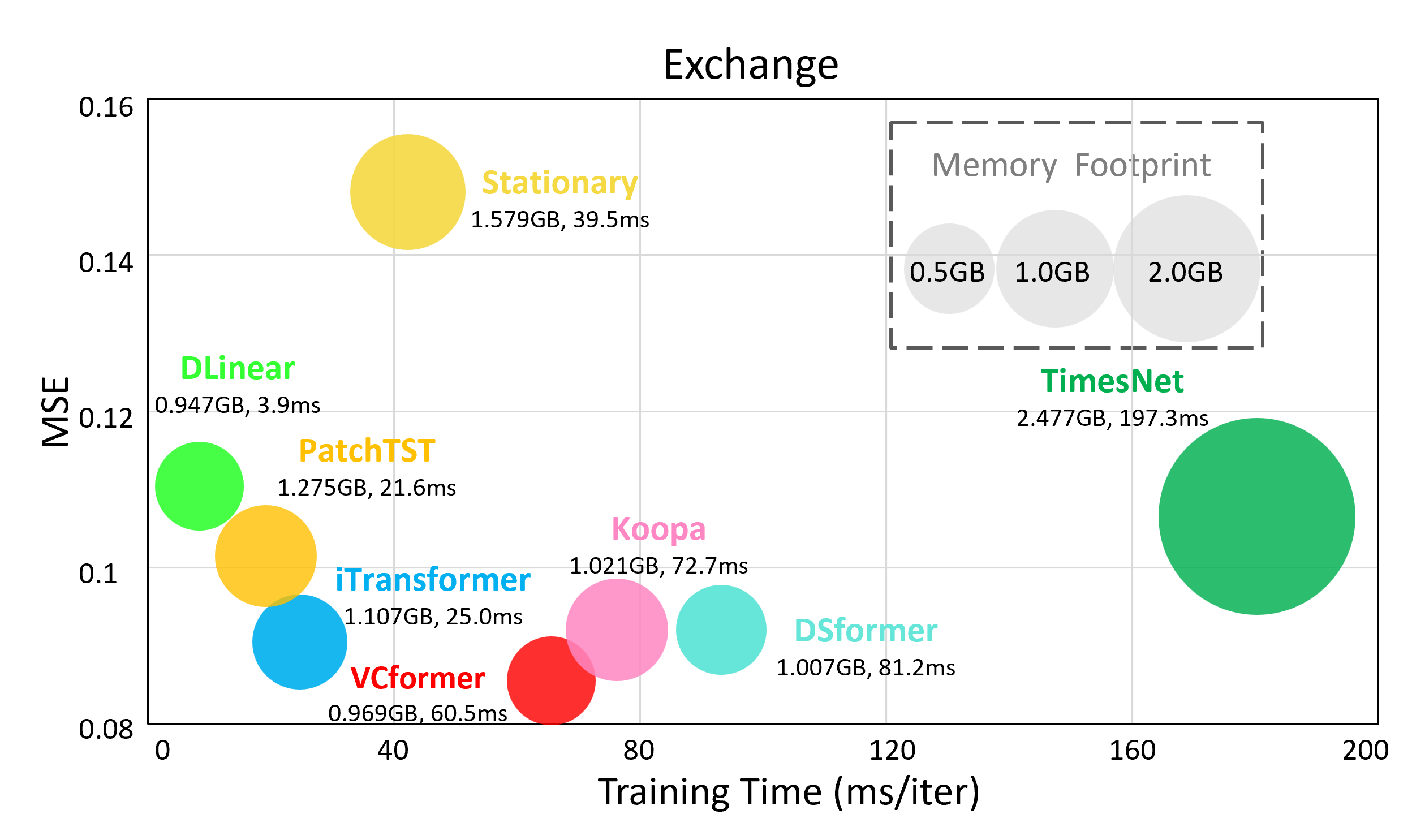}
    \label{fig:exchange}
}
\hfill
\subfloat{
        \includegraphics[width=0.49\textwidth]{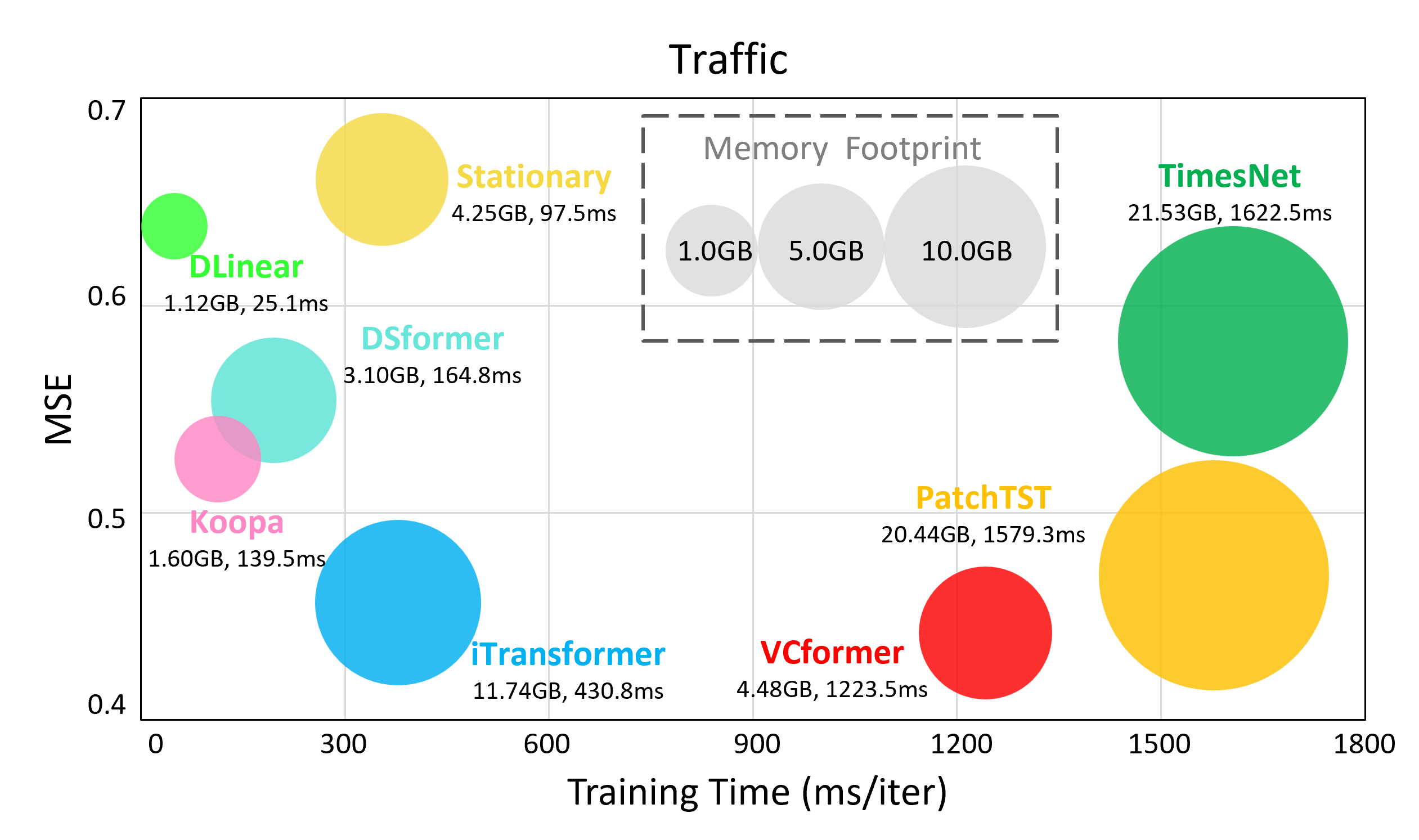}
        \label{fig:traffic}
}
\caption{Model efficiency comparison. The running efficiency of eight models on the Exchange (left) and Traffic (right) dataset with the prediction length $H=96$ and the batch size $B=16$.} 
\label{fig:model_efficiency}
\end{figure*}

\subsection{Model Analysis}

\subsubsection{Ablation Study}
In order to comprehensively understand the individual contributions of the key components in VCformer, we conduct ablation experiments covering experiments with both replacing components (Replace) and removing components (w/o), as shown in Table \ref{tab:ablation_study}. 
From these results, we can conclude that both VCA and KTD are indispensable for the best performance of VCformer, which utilizes lag-correlation on variate dimension and Koopman detector on temporal dimension. After replacing or removing either one of them, the MSE/MAE will increase, which validates their effectiveness. Notably, in the datasets with a large number of variates, specifically Traffic and Electricity, the replacement or removal of the VCA module incurs a remarkable increase in MSE/MAE, i.e., an averaged increase of 9\% over the increase by KTD. 
This suggests that VCA plays a more critical role when dealing with a larger number of variates. On the other hand, within the Exchange dataset noted for its volatility, the substitution or removal of the KTD module, which can capture series non-stationarity, results in more noticeable performance drops than that of the VCA module. 
Conversely, pertaining to the Weather dataset, the experimental results indicate that the omission or replacement of either module doesn't lead to major variances in performance.
These results show that the KTD module is good at learning features from non-stationary time series.

\subsubsection{Multivariate Correlation Analysis}

To enhance the interpretability of learned multivariate correlations by VCA, we provide a visualization case with random selection on series from Exchange. As demonstrated in Figure \ref{fig:learned_map}, the upper part presents the variate correlations inherent within the raw series including both input and prediction sequences. These correlations are calculated by Pearson Correlation coefficient as the following equation:
\begin{equation}
    r_{xy} = \frac{\sum_{i=1}^{L} (x_i - \bar{x})(y_i - \bar{y})}{\sqrt{\sum_{i=1}^{L} (x_i - \bar{x})^2} \cdot \sqrt{\sum_{i=1}^{L} (y_i - \bar{y})^2}}
\end{equation}
where $x_i, y_i \in \mathbb{R}$ run through all time points of the paired variates to be correlated. The lower part portrays the pre-Softmax score maps learned by VCA in both the first and the last layers.

From the red box in Figure \ref{fig:learned_map}, we can observe that the multivariate correlations learned by shallow layer of VCA are more similar to correlations of the input raw series. And as we explore at a deeper layer of VCA, we find that the learned multivariate correlations are closer to the forecasting sequences. This observation indicates that the focus of learned multivariate correlations shifts progressively from input series to prediction sequences. It also enhances interpretability of VCA which aggregates different lag-correlations to represent these variable relationships. 

\subsubsection{Model Efficiency Analysis}

As shown in the Figure \ref{fig:model_efficiency}, we conduct a comparative study of the VCformer's efficiency with seven baselines. Our assessment considers three aspects: training speed, memory consumption and prediction performance.
It can be observed that the time complexity of VCformer is $\mathcal{O}\left(N^2 T \log (T)\right)$. However, the coefficient $N^2$ does not significantly impact the training time when handling datasets with a small number of variates like Exchange. Notably, for the Traffic dataset which contains a large number of variates, the actual computational consumption is not as large as expected. It even outperforms PatchTST and TimesNet, which can be largely attributed to all the required calculations based on matrix operations. And these operations are well parallelized in built-in library.

\section{Conclusion}
In this paper, we address the limitations of the conventional dot product attention mechanism in extracting multivariate correlations. Then we propose VCformer which contains two effective modules. The VCA module can not only mine the lagged cross-correlation implicit in MTS, but also seamlessly integrate into other Transformer-based models. The KTD module employs MLP modules to derive Koopman embeddings and generates Koopman operator to enhance the capability for capturing non-stationarity in MTS. Extensive experiments shows that VCformer achieves SOTA forecasting performance and its VCA module is general enough to improve performance of various Transformer-based models.

\appendix

\section*{Ethical Statement}

There are no ethical issues.

\section*{Acknowledgements}

This work was supported by the National Natural Science Foundation of China under Grants U2013201 and 62203308.

\bibliographystyle{named}
\bibliography{ijcai24}

\newpage
\appendix

\begin{algorithm*}

\caption{VCformer - Overall Architecture}

\begin{algorithmic}[1] 
\setlength{\itemsep}{5pt}
\Require  Input past time series $\mathbf{X}_{in} \in \mathbb{R}^{T \times N}$; Input length $T$; Prediction length $H$; Number of variates $N$; VCformer encoder block number $L$; Token dimension $D$; Koopman embedding dimension $M$; Koopman segment length $S$.
\State $\mathbf{X}_{in}=\mathbf{X}_{in}.\mathrm{transpose}$ \algorithmiccomment{$ \mathbf{X} \in \mathbb{R}^{N \times T}$}
\State  $\triangleright$ Multi-layer Perceptron works on the last dimension to embed series into variate tokens.
\State $\mathbf{X}_{en}^0=\mathrm{MLP}(\mathbf{X}_{in})$ \algorithmiccomment{$\mathbf{X}_{en}^0 \in \mathbb{R}^{N \times D}$}
\For{$l$  \textbf{in} \{1, \ldots, L\}} \algorithmiccomment{Run through VCformer blocks.}
   \State  $\triangleright$ VCA layer is applied on variate tokens to learn multivariate correlations.
    \State $\mathbf{X}_{en}^{l,1}=\mathrm{LayerNorm}(\mathbf{X}_{en}^{l-1}+\mathrm{VCA}(\mathbf{X}_{en}^{l-1}))$ \algorithmiccomment{$\mathbf{X}_{en}^{l,1} \in \mathbb{R}^{N \times D}$}
    \State  $\triangleright$ KTD layer is applied on temporal tokens to capture non-stationarity in time series.
    \State $\mathbf{X}_{en}^{l,2}=\mathrm{LayerNorm}(\mathbf{X}_{en}^{l,1}+\mathrm{KTD}(\mathbf{X}_{en}^{l,1}))$ \algorithmiccomment{$\mathbf{X}_{en}^{l,1} \in \mathbb{R}^{N \times H}$}
    \State  $\triangleright$ LayerNorm is adopted on series representations to reduce variates discrepancies.
    \State $\mathbf{X}_{en}^{l}=\mathbf{X}_{en}^{l,2}$ \algorithmiccomment{$\mathbf{X}_{en}^{l} \in \mathbb{R}^{N \times H}$}
\EndFor
\State $\hat{\mathbf{Y}}=\mathrm{MLP}(X_{en}^{L})$ \algorithmiccomment{Project tokens from the output of Encoder,  $\hat{\mathbf{Y}} \in \mathbb{R}^{N \times H}$}
\State $ \hat{\mathbf{Y}}=\hat{\mathbf{Y}}.\mathrm{transpose}$ \algorithmiccomment{$\mathbf{Y} \in \mathbb{R}^{H \times N}$} 
\State \Return $\hat{\mathbf{Y}}$ \algorithmiccomment{Return the prediction result $\hat{\mathbf{Y}}$}
\end{algorithmic}
\label{algorithm}
\end{algorithm*}

\section{Experiment Details}

\subsection{Datasets}
We conduct extensive experiments on eight real-world datasets including Weather, Electricity, Traffic, Exchange and four ETT datasets (ETTh1, ETTh2, ETTm1, ETTm2). These datasets are widely used in time series forecasting. Here are more details about various datasets as follows:
\begin{itemize}
    \item \textbf{Weather} \footnote{\url{https://www.bgc-jena.mpg.de/wetter/}} contains 21 meteorological measurements such as temperature, precipitation, wind speed and humidity, which is recorded every 10 minutes in German for 2020 whole year.
    \item \textbf{Electricity} \footnote{\url{https://archive.ics.uci.edu/ml/datasets/ElectricityLoadDiagrams20112014}} records hourly electricity consumption data for 321 clients from 2012 to 2014.
    \item \textbf{Traffic} \footnote{\url{https://pems.dot.ca.gov/}} encompasses 862 traffic-related measurements such as vehicle counts, speed and congestion levels collected from sensors or cameras on road networks, which spans from 2015 to 2016 in the San Francisco Bay area.
    \item \textbf{Exchange} \footnote{\url{https://github.com/laiguokun/multivariate-time-series-data}} collects the panel data of daily exchange rates from 8 countries from 1990 to 2016.
    \item \textbf{ETT} \footnote{\url{https://github.com/zhouhaoyi/ETDataset}} (Electricity Transformer Temperature) consists of two years of data from two separate counties in China, with subsets created for different levels of granularity on Time series forecasting problem, including ETTh1 and ETTh2 for 1-hour-level data and ETTm1 for 15-minute-level data.
\end{itemize}
Table \ref{tab:dataset_desc} includes detailed statistics of these datasets. \textit{Timesteps} denotes the total number of time points in dataset. \textit{Sample Frequency} denotes the sampling interval of time points. \textit{Dimension} denotes the number of variates included in dataset.

\subsection{Baselines}

The descriptions of selected baseline methods is given as follows:

\begin{itemize}
    \item iTransformer \cite{iTransformer} inverts the vanilla Transformer backbone for time series forecasting, applying the self-attention mechanism on learning multivariate correlations and encoding series representations by FFN. The source code is available at \url{https://github.com/thuml/iTransformer}.
    \item DSformer \cite{DSformer}  proposes double sampling (DS) block and the temporal variable attention (TVA) block to mine the global and local information as well as variable correlations. The source code is available at \url{https://github.com/ChengqingYu/DSformer}.
    \item PatchTST \cite{PatchTST} is a Transformer-based approach for MTS forecasting that utilizes patching and channel-independence strategies. The source code is available at \url{https://github.com/yuqinie98/patchtst}.
    \item Crossformer \cite{Crossformer} is the first Transformer-based model that employed the two-stage attention to capture both cross-dim and cross-time dependencies.The source code is available at \url{https://github.com/Thinklab-SJTU/Crossformer}.
    \item Koopa \cite{Koopa} is a MLP-based model that disentangle time series into time-invariant and time-variant dynamics. The source code is available at \url{https://github.com/thuml/Koopa}.
    \item TimesNet \cite{TimesNet} transforms 1D time series into 2D tensors based on Fourier Transform and applies inception blocks on 2D tensor to capture intricate temporal variations.
    The source code is available at \url{https://github.com/thuml/TimesNet}.
    \item DLinear \cite{DLinear} questions the efficiency of Transformer-based forecasters and leverage a simple one-layer linear model achieving superior performance on multiple datasets. The source code is available at \url{https://github.com/curelab/LTSF-Linear}.
    \item Nonstationary Transformer \cite{NSTrans} focuses on the over-stationarization problem, designing the De-stationary attention mechanism to tackle it. The source code is available at \url{https://github.com/thuml/Nonstationary_Transformers}.
\end{itemize}

\begin{table}[]
    \centering
    \begin{tabular}{c|c|c|c}
         \toprule
         Datasets & Timesteps & Sample Frequency & Dimension \\
         \midrule
         Weather & 52696 & 10 min & 21 \\
         Electricity & 26304 & 1 hour & 321 \\
         Traffic & 17544 & 1 hour & 862 \\
         Exchange & 7207 & 1 day & 8 \\
         ETTh1 & 17420 & 1 hour & 7 \\
         ETTh2 & 17420 & 1 hour & 7 \\
         ETTm1 & 69680 & 15 min & 7 \\
         ETTm2 & 69680 & 15 min & 7 \\
         \bottomrule
    \end{tabular}
    \caption{The detail statistics of datasets}
    \label{tab:dataset_desc}
\end{table}

\subsection{Experiment Setting and Hyperparameter}

Our experiments, except for the Traffic dataset, are conducted with Pytorch 1.11.0 on single NVIDIA Tesla P100 GPU, which is equipped with 16GB CUDA memory. For the Traffic dataset, we run the experiments on multiple GPUs to facilitate the process. Following \cite{TimesNet}, we set lookback window length $T=96$ for all the experiments to ensure fairness and various prediction horizons $H \in \left\{96, 192, 336, 720\right\}$. All the experiments are trained by Adam using L2 loss and repeated three times to avoid accidents. 
All the baselines are reproduced by their official implementations with recommended hyperparameters. The batch size is set to 16 for most baselines, while Crossformer is set to 8 due to its large memory consumption. The learning rate is initialized to 0.5 and decays exponentially as training epochs grow. We set the number of encoder layers $L \in \left\{1,2,3\right\}$, the dimension of Koopman embedding $M \in \left\{256, 512, 1024\right\}$, the projection space of inverted embedding $D \in \left\{128,256,512\right\}$ and the length of Koopman segment $S=32$. And we also provide the pseudo-code of VCformer in Algorithm \ref{algorithm}.

\section{Proof of Efficient Computation}
Given series $\{\mathbf{k}_j\}$ that lags behind $\{\mathbf{q}_i\}$ by $\tau$ steps, we denote $\mathcal{K}_f$ and $\mathcal{Q}_f$ as the respective frequency components. According to stochastic process theory, the cross-correlation between them can be defined by:

$$
R_{\mathbf{q}_i, \mathbf{k}_j}(\tau) =\frac{1}{T} \sum_{t=1}^T \mathbf{k}_j(t-\tau) \mathbf{q}_i(t)
$$
With $\mathbf{k}_j(t-\tau)$ denoted as $\check{\mathbf{k}_j}(\tau-t)$ and $(\tau-t)$ denoted as $\tau^{\prime}$, we derive the Fourier Transform of $\left\{\sum_{t=1}^{T} \mathbf{q}_i(t) \check{\mathbf{k}_j}(\tau-t)\right\}_{\tau=1}^{T}$ as follows:

$$
\begin{aligned}
& \mathcal{F}\left(\sum_{t=1}^{T} \mathbf{q}_i(t) \check{\mathbf{k}_j}(\tau-t)\right)_f \\ &=\sum_{\tau=1}^{T}\left(\sum_{t=1}^{T} \mathbf{q}_i(t) \check{\mathbf{k}_j}(\tau-t)\right) e^{-i 2 \pi f \tau} \\
&= \sum_{\tau=1}^{T}\mathbf{q}_i(t) \left(\sum_{t=1}^{T} \check{\mathbf{k}_j}(\tau-t) e^{-i 2 \pi f \tau}\right) \\
&= \sum_{\tau=1}^{T}\mathbf{q}_i(t) e^{-i 2 \pi f t}\left(\sum_{t=1}^{T} \check{\mathbf{k}_j}(\tau-t) e^{-i 2 \pi f (\tau-t)}\right) \\
& =\mathcal{Q}_f\left(\sum_{\tau=1}^{T} \check{\mathbf{k}_j}\left(\tau^{\prime}\right) e^{-i 2 \pi f \tau^{\prime}}\right) .
\end{aligned}
$$
Assuming $\check{\mathbf{k}_j}$ is $T$-periodic, we have

$$
\begin{aligned}
& \sum_{\tau^{\prime}=1-t}^{T-t} \check{\mathbf{k}_j}\left(\tau^{\prime}\right) e^{-i 2 \pi f \tau^{\prime}} \\
&= \sum_{\tau^{\prime}=1-t}^{0} \check{\mathbf{k}_j}\left(\tau^{\prime}\right) \cdot e^{-i 2 \pi  f\left(\tau^{\prime}+T\right)}+\sum_{\tau^{\prime}=1}^{T-t} \check{\mathbf{k}_j}\left(\tau^{\prime}\right) \cdot e^{-i 2 \pi f \tau^{\prime}} \\
& =\sum_{\tau^{\prime}=1-t}^{0} \check{\mathbf{k}_j}\left(\tau^{\prime}+T\right) \cdot e^{-i 2 \pi  f\left(\tau^{\prime}+T\right)} + \sum_{\tau^{\prime}=1}^{T-t} \check{\mathbf{k}_j}\left(\tau^{\prime}\right) \cdot e^{-i 2 \pi f \tau^{\prime}} \\
& =\sum_{\tau^{\prime}=T-t+1}^{T} \check{\mathbf{k}_j}\left(\tau^{\prime}\right) \cdot e^{-i 2 \pi  f\tau^{\prime}} + \sum_{\tau^{\prime}=1}^{T-t} \check{\mathbf{k}_j}\left(\tau^{\prime}\right) \cdot e^{-i 2 \pi f \tau^{\prime}} \\
& =\sum_{\tau^{\prime}=1}^{T} \check{\mathbf{k}_j}\left(\tau^{\prime}\right) \cdot e^{-i 2 \pi  f\tau^{\prime}} \\
& =\mathcal{F}(\check{\mathbf{k}_j})_f . \\
\end{aligned}
$$
Due to the conjugate symmetry of the DFT on real-valued signals, we have
$$
\mathcal{F}(\check{\mathbf{k}_j})_f=\overline{\mathcal{F}(\mathbf{k}_j)}_f=\overline{\mathcal{K}_f},
$$
where the bar is the conjugate operation. Thereby, we can obtain

$$
\mathcal{F}\left(\sum_{t=1}^{T} \mathbf{q}_i(t) \check{\mathbf{k}_j}(\tau-t)\right)_f=\mathcal{Q}_f \overline{K_f}
$$
Finally, we can obtain $R_{\mathbf{q}_i, \mathbf{k}_j}(\tau)$ as:

$$
R_{\mathbf{q}_i, \mathbf{k}_j}(\tau) = \frac{1}{T} \mathcal{F}^{-1}\left(\mathcal{F}\left(\mathbf{q}_i\right) \odot \overline{\mathcal{F}\left(\mathbf{k}_j\right)}\right)_\tau
$$
Note that $R_{\mathbf{q}_i, \mathbf{k}_j}(\tau) \in [-1,1]$ when $\mathbf{q}_i$ and $\mathbf{k}_j$ have been normalized.

\section{Hyperparameter Sensitivity}
\begin{figure*}
    \centering 
    \resizebox{0.95\textwidth}{!}{\includegraphics{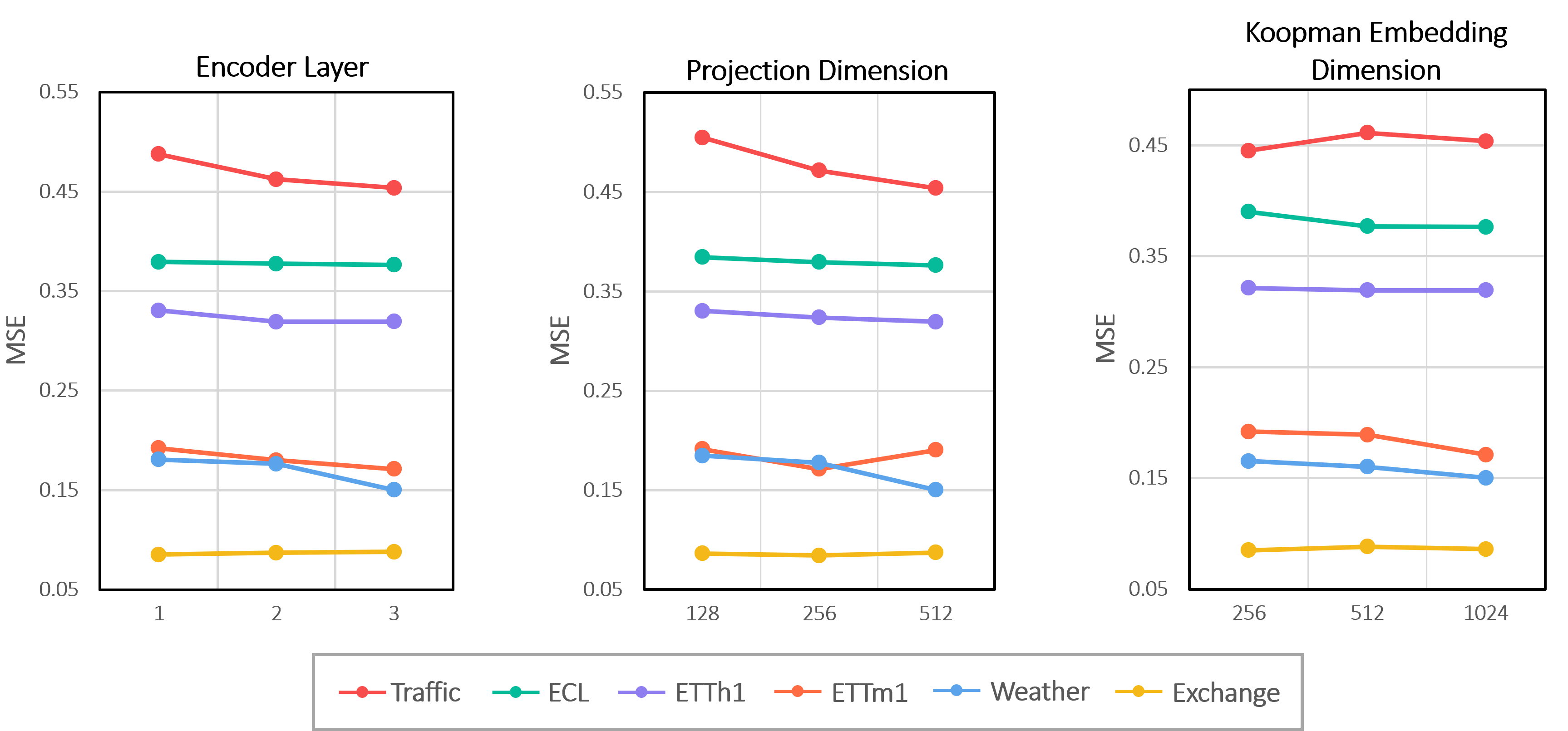}}
    \caption{Hyperparameter sensitivity with respect to the encoder layer, the Koopman embedding dimension and the projection dimension of variate tokens.}
    \label{fig:hyperparm_sensitivity}
\end{figure*}

We conduct experiments about the hyperparameter sensitivity of VCformer as shown in Figure \ref{fig:hyperparm_sensitivity}, which include three factors: the number of encoder layers $L$, the dimension  $D$ of inverted embeddings and the dimension $M$ of Koopman embeddings. Based on the experimental results, we find that as the hyperparameter values increasing, the performance on most datasets will have an improvement except for Exchange dataset. It can be attributed to the overfitting problem which is caused by the high volatility of Exchange dataset and the increasing parameters of model. Moreover, compared to other datasets, the Electricity and ETT datasets exhibit low sensitivity to changes in the hyperparameter. 
 
\section{Full Results}
\subsection{Full Ablation Results}
Due to the limited pages, we list the overall ablation study results on the effect of VCA and KTD in VCformer as shown in Table \ref{tab:full_ablation}. The detailed ablations contain two type of experiments denoted as removing components (w/o) and replacing components (replace). 

In Table \ref{tab:full_ablation}, among different architecture designs, VCformer utilizes the lagged-correlation inherent between variates by the VCA module and captures the non-stationarity in time series by the KTD module. It thus exhibits top-tier performance (with the average results in bold). Specifically, the replacement of VCA achieves inferior performance which indicates the deficiency of vanilla point-wise self-attention mechanism on learning the multivariate correlations. It can be attributed to the neglect of existence of lagged-correlations. With the increasing number of variates, the deterioration in performance becomes increasingly evident. It implies that the importance of capturing multivariate correlations is ever more highlighted. Besides, the replacement of KTD by FFN also gets the worse performance especially on the Exchange dataset which is noted for the non-stationarity. This phenomenon indicates the effectiveness of KTD module for the capability to mine the complex temporal dependencies. 

\subsection{Full VCA Generality Results}
In Table \ref{tab:full_generality}, we apply the VCA module to six Transformer-based models and set the better average results in bold. "Replace" denotes that the VCA module is used to substitute the self-attention mechanism in these Transformer variants. From the results, we can find that the number of bold average results for "Replace" (count=52) is much more than "Origin" (count=8). Due to the capability to learn the multivariate correlations, the replacements by VCA module significantly improves the performance of these Transformer-based methods.
\section{Visualization of MTS Forecasting}

\begin{figure*}[t]
    \centering 
    \resizebox{\textwidth}{!}{\includegraphics{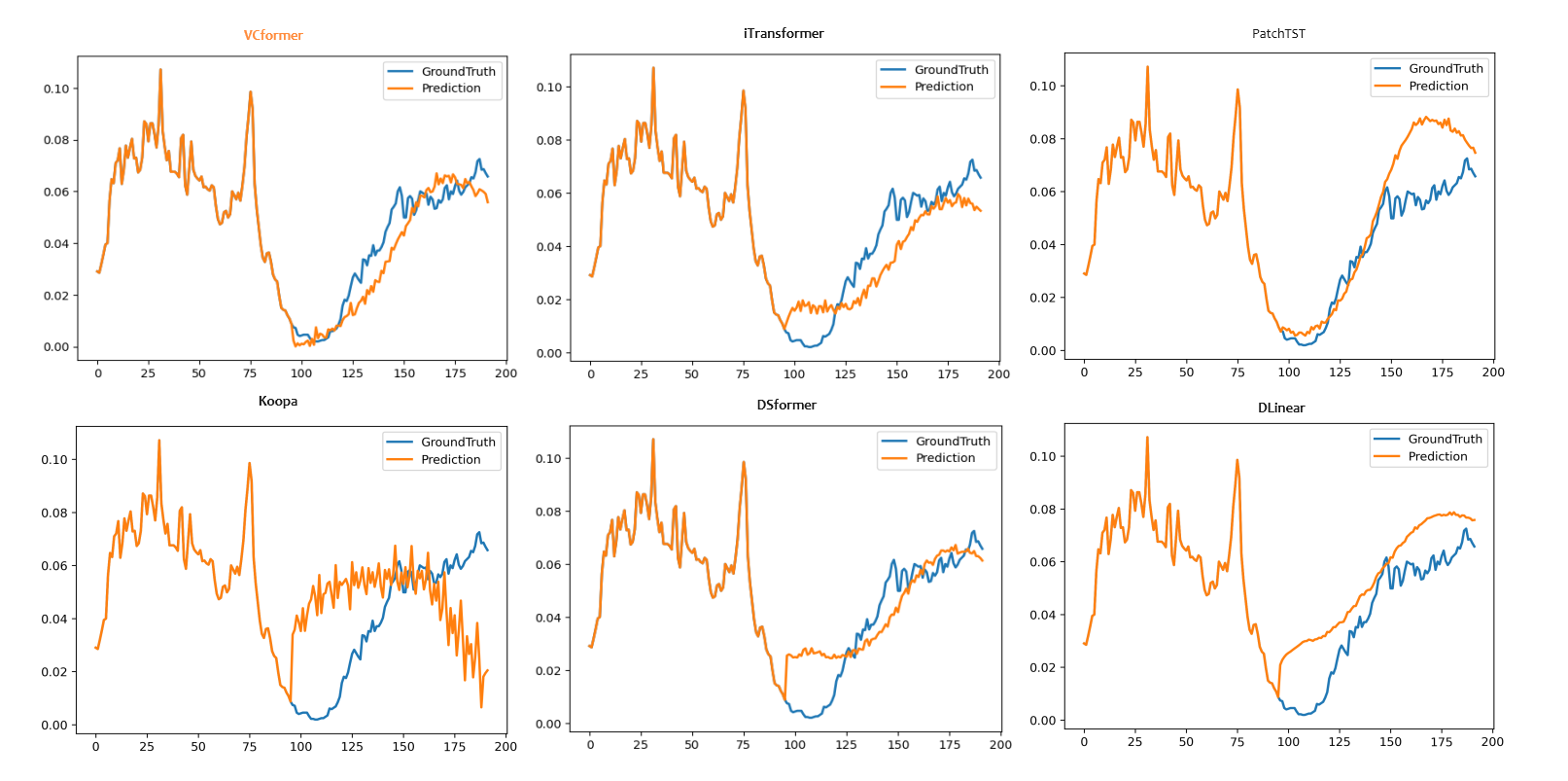}}
    \caption{Visualization of input-96-predict-96 results on the Weather dataset.}
    \label{fig:weather_show}
\end{figure*}

\begin{figure*}[t]
    \centering 
    \resizebox{\textwidth}{!}{\includegraphics{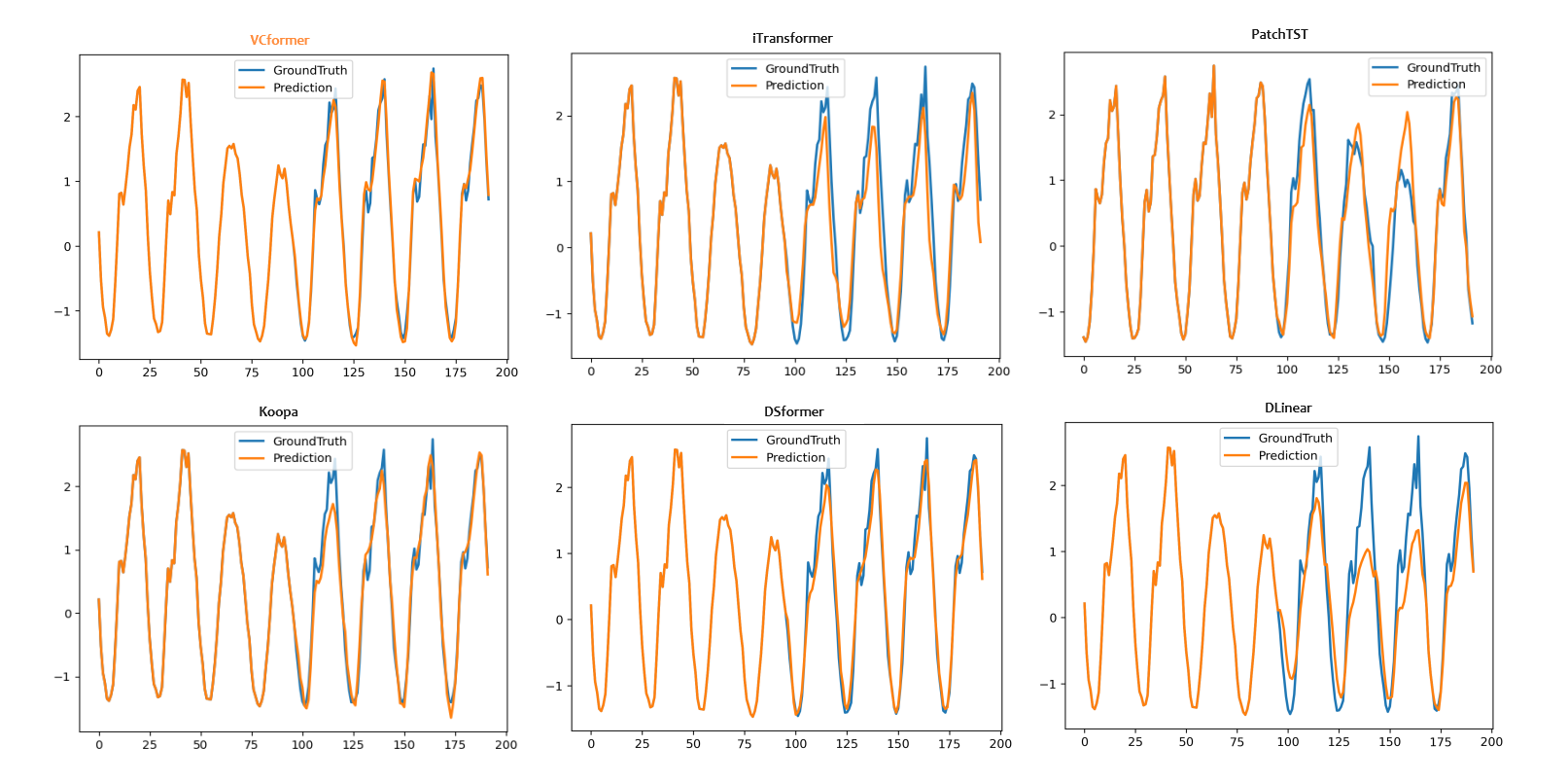}}
    \caption{Visualization of input-96-predict-96 results on the Traffic dataset.}
    \label{fig:traffic_show}
\end{figure*}

\begin{figure*}[t]
    \centering 
    \resizebox{\textwidth}{!}{\includegraphics{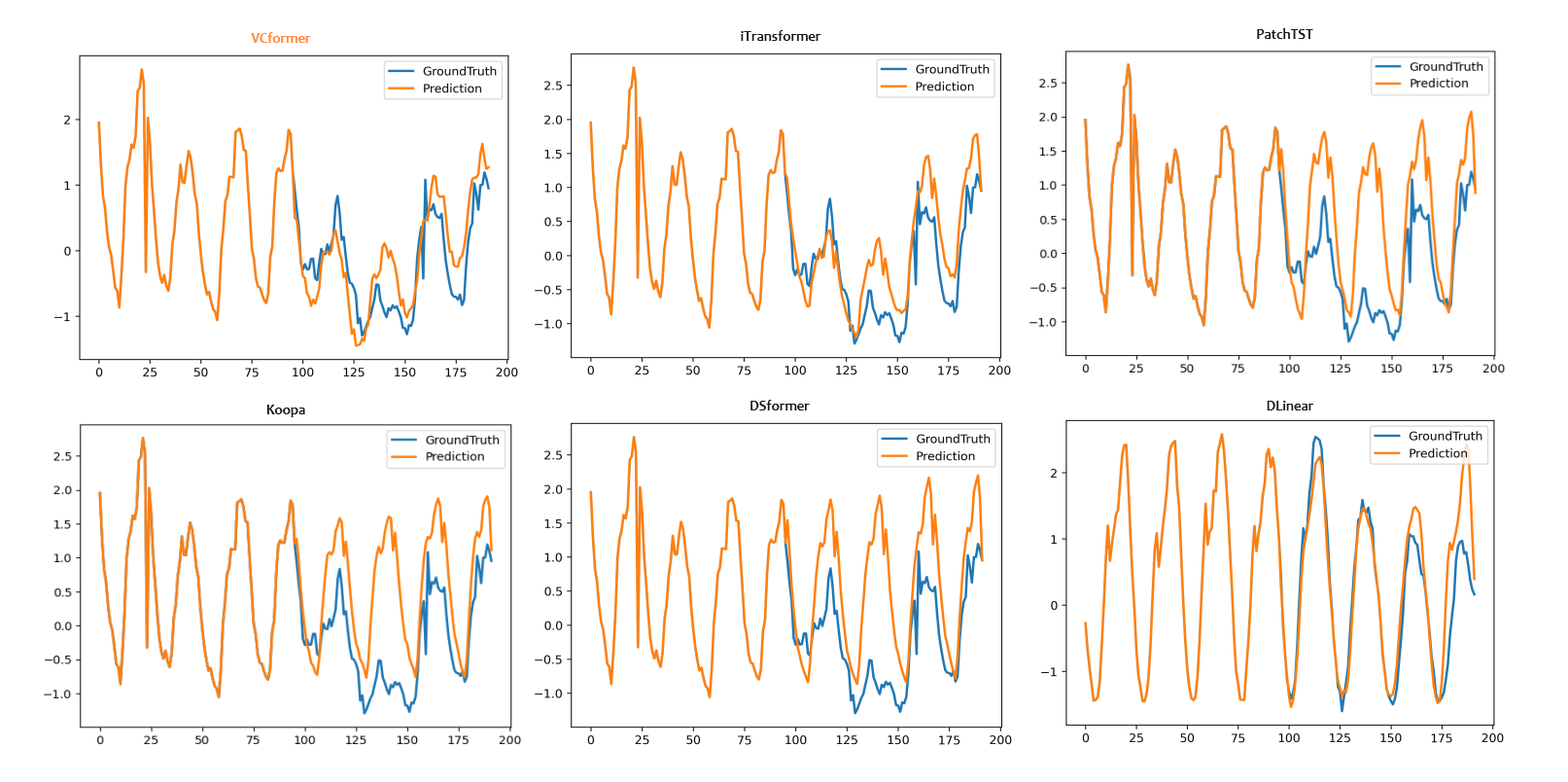}}
    \caption{Visualization of input-96-predict-96 results on the Electricity dataset.}
    \label{fig:ECL_show}
\end{figure*}

To illustrate the prediction performance of VCformer more intuitively, we list several prediction showcases of three datasets in Figure \ref{fig:weather_show}-\ref{fig:ECL_show} given by VCformer, iTransformer \cite{iTransformer}, DSformer \cite{DSformer}, PatchTST \cite{PatchTST}, DLinear \cite{DLinear} and Koopa \cite{Koopa}. The look-back window and prediction length are both set to 96 for all models. From the visualization, VCformer exhibit precise prediction to the ground truth and thus achieve superior performance.

\begin{table*}[t]
    \centering
    \resizebox{\textwidth}{!}{
        \begin{tabular}{c|c|c|c|cc|cc|cc|cc|cc}
        \toprule
        \multirow{2}{*}{Design}& \multirow{2}{*}{Variate} & \multirow{2}{*}{Temporal} & Prediction 
        & \multicolumn{2}{c}{Exchange} & \multicolumn{2}{c}{Traffic} & \multicolumn{2}{c}{Electricity} & \multicolumn{2}{c}{Weather} & \multicolumn{2}{c}{ETTh2}\\
        \cmidrule{4-14}
        & & & Lengths & MSE & MAE & MSE & MAE & MSE & MAE & MSE & MAE & MSE & MAE\\
        \midrule
        \multirow{5}{*}{VCformer} & \multirow{5}{*}{VCA} & \multirow{5}{*}{KTD} 
            & 96  & 0.085 & 0.205 & 0.454 & 0.310 & 0.150 & 0.242 & 0.171 & 0.220 & 0.292 & 0.344\\
        & & & 192 & 0.176 & 0.195 & 0.468 & 0.315 & 0.167 & 0.255 & 0.230 & 0.266 & 0.377 & 0.396\\
        & & & 336 & 0.328 & 0.328 & 0.486 & 0.325 & 0.182 & 0.270 & 0.280 & 0.299 & 0.417 & 0.430\\
        & & & 720 & 0.850 & 0.914 & 0.524 & 0.348 & 0.221 & 0.302 & 0.352 & 0.344 & 0.423 & 0.443\\
        \cmidrule{4-14}
        & &  & Avg & \textbf{0.360} & \textbf{0.402} & \textbf{0.483} & \textbf{0.325} & \textbf{0.180} & \textbf{0.267} & \textbf{0.258} & \textbf{0.282} & 0.377 & 0.403\\
        \midrule
        \multirow{10}{*}{Replace}
        & \multirow{5}{*}{Attention} & \multirow{5}{*}{KTD}
            & 96  & 0.100 & 0.235 & 0.490 & 0.235 & 0.162 & 0.259 & 0.173 & 0.218 & 0.292 & 0.345 \\
        & & & 192 & 0.195 & 0.331 & 0.513 & 0.331 & 0.180 & 0.275 & 0.235 & 0.273 & 0.376 & 0.395 \\
        & & & 336 & 0.350 & 0.485 & 0.529 & 0.485 & 0.203 & 0.294 & 0.290 & 0.301 & 0.419 & 0.432 \\
        & & & 720 & 0.914 & 0.735 & 0.577 & 0.735 & 0.245 & 0.330 & 0.363 & 0.349 & 0.422 & 0.441 \\
        \cmidrule{4-14}
        & & & Avg & 0.390 & 0.447 & 0.527 & 0.351 & 0.198 & 0.290 & 0.265 & 0.285 & \textbf{0.376} & \textbf{0.401} \\
        \cmidrule{2-14}
        & \multirow{5}{*}{VCA} & \multirow{5}{*}{FFN}
            & 96  & 0.110 & 0.265 & 0.469 & 0.321 & 0.155 & 0.246 & 0.175 & 0.223 & 0.295 & 0.346 \\
        & & & 192 & 0.219 & 0.377 & 0.481 & 0.327 & 0.170 & 0.259 & 0.234 & 0.270 & 0.381 & 0.401 \\
        & & & 336 & 0.389 & 0.500 & 0.490 & 0.340 & 0.186 & 0.276 & 0.289 & 0.304 & 0.423 & 0.436 \\
        & & & 720 & 0.982 & 0.779 & 0.551 & 0.359 & 0.226 & 0.310 & 0.357 & 0.351 & 0.429 & 0.452 \\
        \cmidrule{4-14}
        & & & Avg & 0.425 & 0.480 & 0.498 & 0.337 & 0.184 & 0.273 & 0.264 & 0.287 & 0.382 & 0.409 \\
        \midrule
        \multirow{10}{*}{w/o}
        & \multirow{5}{*}{w/o} & \multirow{5}{*}{KTD}
            & 96  & 0.092 & 0.216 & 0.501 & 0.342 & 0.165 & 0.262 & 0.177 & 0.225 & 0.301 & 0.357 \\
        & & & 192 & 0.201 & 0.356 & 0.520 & 0.358 & 0.179 & 0.272 & 0.236 & 0.271 & 0.385 & 0.406 \\
        & & & 336 & 0.357 & 0.493 & 0.535 & 0.371 & 0.199 & 0.290 & 0.291 & 0.302 & 0.433 & 0.448 \\
        & & & 720 & 1.025 & 0.751 & 0.596 & 0.389 & 0.247 & 0.329 & 0.381 & 0.362 & 0.457 & 0.469 \\
        \cmidrule{4-14}
        & & & Avg & 0.419 & 0.454 & 0.538 & 0.365 & 0.198 & 0.288 & 0.271 & 0.290 & 0.394 & 0.420 \\
        \cmidrule{2-14}
        & \multirow{5}{*}{VCA} & \multirow{5}{*}{w/o}
            & 96  & 0.188 & 0.270 & 0.501 & 0.333 & 0.160 & 0.253 & 0.176 & 0.225 & 0.305 & 0.360 \\
        & & & 192 & 0.215 & 0.389 & 0.520 & 0.341 & 0.174 & 0.261 & 0.235 & 0.273 & 0.387 & 0.403 \\
        & & & 336 & 0.403 & 0.573 & 0.512 & 0.356 & 0.192 & 0.283 & 0.293 & 0.309 & 0.436 & 0.447 \\
        & & & 720 & 1.025 & 0.792 & 0.577 & 0.375 & 0.234 & 0.323 & 0.371 & 0.357 & 0.461 & 0.470 \\
        \cmidrule{4-14}
        & & & Avg & 0.440 & 0.506 & 0.518 & 0.351 & 0.190 & 0.280 & 0.269 & 0.291 & 0.397 & 0.420 \\
        \midrule
    \end{tabular}
    }
    \caption{Full results of the ablation on VCformer. We conduct substitution and removal experiments on two key components (VCA \& KTD) of VCformer respectively on the dimensions they represent (Variate \& Temporal).}
    \label{tab:full_ablation}
\end{table*}

\begin{table*}[t]
    \centering
    \resizebox{\textwidth}{!}{
    
        \begin{tabular}{c|c|c|cc|cc|cc|cc|cc|cc}
        \toprule
        \multicolumn{3}{c}{Models} & \multicolumn{2}{c}{iTransformer} & \multicolumn{2}{c}{DSformer} & \multicolumn{2}{c}{Crossformer} & \multicolumn{2}{c}{Stationary} & \multicolumn{2}{c}{Autoformer}  & \multicolumn{2}{c}{Informer} \\
        \midrule
        \multicolumn{3}{c}{Metric} & MSE & MAE & MSE & MAE & MSE & MAE & MSE & MAE & MSE & MAE & MSE & MAE \\
        \midrule 
        \multirow{10}{*}{Electricity} & \multirow{5}{*}{Original} 
          & 96  & 0.154 & 0.245 & 0.164 & 0.261 & 0.153 & 0.250 & 0.172 & 0.275 & 0.201 & 0.317 & 0.274 & 0.368 \\
        & & 192 & 0.169 & 0.258 & 0.177 & 0.272 & 0.223 & 0.329 & 0.187 & 0.287 & 0.222 & 0.334 & 0.296 & 0.386 \\
        & & 336 & 0.185 & 0.275 & 0.201 & 0.294 & 0.191 & 0.291 & 0.208 & 0.307 & 0.231 & 0.338 & 0.300 & 0.394 \\
        & & 720 & 0.225 & 0.308 & 0.242 & 0.327 & 0.609 & 0.568 & 0.235 & 0.329 & 0.254 & 0.361 & 0.373 & 0.439 \\
        \cmidrule{3-15}
        & & Avg & 0.183 & 0.272 & 0.196 & 0.288 & 0.294 & 0.359 & 0.200 & 0.299 & 0.227 & 0.338 & 0.311 & 0.397 \\
        \cmidrule{2-15}
        & \multirow{5}{*}{Replace} 
          & 96  & 0.151 & 0.243 & 0.159 & 0.252 & 0.153 & 0.251 & 0.163 & 0.262 & 0.170 & 0.273 & 0.195 & 0.301 \\
        & & 192 & 0.168 & 0.256 & 0.172 & 0.263 & 0.219 & 0.328 & 0.175 & 0.270 & 0.195 & 0.290 & 0.210 & 0.315 \\
        & & 336 & 0.183 & 0.272 & 0.190 & 0.285 & 0.190 & 0.288 & 0.195 & 0.286 & 0.200 & 0.295 & 0.231 & 0.339 \\
        & & 720 & 0.223 & 0.305 & 0.235 & 0.315 & 0.610 & 0.566 & 0.230 & 0.310 & 0.237 & 0.331 & 0.266 & 0.361 \\
        \cmidrule{3-15}
        & & Avg & \textbf{0.181} & \textbf{0.269} & \textbf{0.189} & \textbf{0.279} & \textbf{0.293} & \textbf{0.358} & \textbf{0.191} & \textbf{0.282} & \textbf{0.200} & \textbf{0.297} & \textbf{0.226} & \textbf{0.329} \\
        \midrule
        \multirow{10}{*}{Exchange} & \multirow{5}{*}{Original} 
          & 96  & 0.090 & 0.211 & 0.092 & 0.216 & 0.139 & 0.265 & 0.154 & 0.297 & 0.197 & 0.323 & 0.847 & 0.752 \\
        & & 192 & 0.186 & 0.307 & 0.189 & 0.312 & 0.241 & 0.375 & 0.374 & 0.447 & 0.300 & 0.369 & 1.204 & 0.895 \\
        & & 336 & 0.339 & 0.424 & 0.348 & 0.430 & 0.392 & 0.468 & 0.548 & 0.563 & 0.509 & 0.524 & 1.672 & 1.036 \\
        & & 720 & 0.898 & 0.718 & 0.947 & 0.740 & 1.110 & 0.802 & 0.987 & 0.777 & 1.447 & 0.941 & 2.478 & 1.310 \\
        \cmidrule{3-15}
        & & Avg & 0.378 & 0.415 & \textbf{0.394} & \textbf{0.424} & 0.471 & 0.478 & 0.516 & 0.521 & 0.613 & 0.539 & 1.550 & 0.998 \\
        \cmidrule{2-15}
        & \multirow{5}{*}{Replace} 
          & 96  & 0.088 & 0.207 & 0.095 & 0.217 & 0.097 & 0.225 & 0.100 & 0.235 & 0.124 & 0.278 & 0.301 & 0.414 \\
        & & 192 & 0.183 & 0.302 & 0.192 & 0.320 & 0.197 & 0.332 & 0.220 & 0.301 & 0.255 & 0.323 & 0.441 & 0.615 \\
        & & 336 & 0.334 & 0.420 & 0.349 & 0.435 & 0.350 & 0.447 & 0.405 & 0.479 & 0.443 & 0.501 & 0.573 & 0.729 \\
        & & 720 & 0.866 & 0.695 & 0.960 & 0.745 & 0.973 & 0.755 & 0.860 & 0.844 & 1.501 & 0.893 & 1.109 & 0.883 \\
        \cmidrule{3-15}
        & & Avg & \textbf{0.368} & \textbf{0.406} & 0.399 & 0.429 & \textbf{0.404} & \textbf{0.440} & \textbf{0.396} & \textbf{0.465} & \textbf{0.581} & \textbf{0.499} & \textbf{0.606} & \textbf{0.660} \\
        \midrule
        \multirow{10}{*}{Traffic} & \multirow{5}{*}{Original} 
          & 96  & 0.717 & 0.466 & 0.546 & 0.352 & 0.530 & 0.285 & 0.612 & 0.338 & 0.613 & 0.388 & 0.719 & 0.391 \\
        & & 192 & 0.472 & 0.320 & 0.547 & 0.347 & 0.607 & 0.311 & 0.613 & 0.340 & 0.616 & 0.379 & 0.696 & 0.379 \\
        & & 336 & 0.488 & 0.330 & 0.562 & 0.352 & 0.642 & 0.324 & 0.618 & 0.328 & 0.622 & 0.337 & 0.777 & 0.420 \\
        & & 720 & 0.530 & 0.361 & 0.597 & 0.370 & 0.592 & 0.380 & 0.653 & 0.355 & 0.660 & 0.408 & 0.864 & 0.472 \\
        \cmidrule{3-15}
        & & Avg & 0.552 & 0.369 & 0.563 & 0.355 & 0.593 & 0.325 & 0.624 & 0.340 & 0.628 & 0.378 & 0.764 & 0.416 \\
        \cmidrule{2-15}
        & \multirow{5}{*}{Replace} 
          & 96  & 0.495 & 0.334 & 0.519 & 0.341 & 0.527 & 0.283 & 0.540 & 0.321 & 0.559 & 0.357 & 0.590 & 0.371 \\
        & & 192 & 0.470 & 0.319 & 0.525 & 0.343 & 0.565 & 0.299 & 0.548 & 0.324 & 0.563 & 0.355 & 0.601 & 0.381 \\
        & & 336 & 0.487 & 0.328 & 0.541 & 0.349 & 0.583 & 0.325 & 0.554 & 0.331 & 0.570 & 0.366 & 0.595 & 0.382 \\
        & & 720 & 0.526 & 0.359 & 0.568 & 0.363 & 0.591 & 0.379 & 0.579 & 0.362 & 0.601 & 0.385 & 0.622 & 0.407 \\
        \cmidrule{3-15}
        & & Avg & \textbf{0.495} & \textbf{0.335} & \textbf{0.538} & \textbf{0.349} & \textbf{0.567} & \textbf{0.322} & \textbf{0.555} & \textbf{0.335} & \textbf{0.573} & \textbf{0.366} & \textbf{0.602} & \textbf{0.385} \\

        \midrule
        \multirow{10}{*}{Weather} & \multirow{5}{*}{Original} 
          & 96  & 0.174 & 0.214 & 0.170 & 0.217 & 0.185 & 0.248 & 0.205 & 0.265 & 0.266 & 0.336 & 0.300 & 0.384 \\
        & & 192 & 0.221 & 0.254 & 0.253 & 0.296 & 0.229 & 0.305 & 0.233 & 0.274 & 0.336 & 0.367 & 0.598 & 0.544 \\
        & & 336 & 0.278 & 0.296 & 0.285 & 0.310 & 0.323 & 0.285 & 0.296 & 0.317 & 0.359 & 0.395 & 0.578 & 0.523 \\
        & & 720 & 0.354 & 0.349 & 0.395 & 0.391 & 0.665 & 0.356 & 0.372 & 0.365 & 0.419 & 0.428 & 1.059 & 0.741 \\
        \cmidrule{3-15}
        & & Avg & \textbf{0.257} & \textbf{0.278} & 0.276 & 0.304 & 0.351 & 0.299 & 0.276 & 0.305 & 0.345 & 0.382 & 0.634 & 0.548 \\
        \cmidrule{2-15}
        & \multirow{5}{*}{Replace} 
          & 96  & 0.175 & 0.215 & 0.175 & 0.100 & 0.186 & 0.250 & 0.193 & 0.260 & 0.244 & 0.329 & 0.269 & 0.372 \\
        & & 192 & 0.223 & 0.260 & 0.244 & 0.288 & 0.237 & 0.310 & 0.230 & 0.269 & 0.319 & 0.359 & 0.493 & 0.469 \\
        & & 336 & 0.285 & 0.303 & 0.269 & 0.295 & 0.301 & 0.279 & 0.293 & 0.315 & 0.343 & 0.387 & 0.489 & 0.475 \\
        & & 720 & 0.356 & 0.352 & 0.379 & 0.380 & 0.490 & 0.417 & 0.365 & 0.361 & 0.401 & 0.392 & 0.882 & 0.693 \\
        \cmidrule{3-15}
        & & Avg & 0.260 & 0.283 & \textbf{0.267} & \textbf{0.293} & \textbf{0.304} & \textbf{0.314} & \textbf{0.270} & \textbf{0.301} & \textbf{0.327} & \textbf{0.367} & \textbf{0.500} & \textbf{0.508} \\
        \midrule
        \multirow{10}{*}{ETTh2} & \multirow{5}{*}{Original} 
          & 96  & 0.292 & 0.344 & 0.296 & 0.351 & 0.745 & 0.584 & 0.477 & 0.462 & 0.346 & 0.388 & 3.755 & 1.525 \\
        & & 192 & 0.375 & 0.396 & 0.399 & 0.414 & 0.877 & 0.656 & 0.571 & 0.507 & 0.456 & 0.452 & 5.602 & 1.931 \\
        & & 336 & 0.418 & 0.430 & 0.434 & 0.443 & 1.043 & 0.731 & 0.608 & 0.534 & 0.482 & 0.486 & 4.721 & 1.835 \\
        & & 720 & 0.424 & 0.443 & 0.454 & 0.463 & 1.104 & 0.763 & 0.508 & 0.487 & 0.515 & 0.511 & 3.647 & 1.625 \\
        \cmidrule{3-15}
        & & Avg & \textbf{0.377} & \textbf{0.403} & \textbf{0.396} & \textbf{0.418} & 0.942 & 0.684 & 0.541 & 0.498 & 0.450 & 0.459 & 1.301 & 3.874 \\
        \cmidrule{2-15}
        & \multirow{5}{*}{Replace} 
          & 96  & 0.293 & 0.350 & 0.300 & 0.359 & 0.379 & 0.402 & 0.430 & 0.445 & 0.344 & 0.373 & 1.293 & 0.925 \\
        & & 192 & 0.380 & 0.399 & 0.399 & 0.417 & 0.410 & 0.433 & 0.523 & 0.483 & 0.435 & 0.441 & 1.595 & 0.957 \\
        & & 336 & 0.420 & 0.431 & 0.434 & 0.445 & 0.455 & 0.489 & 0.557 & 0.519 & 0.472 & 0.469 & 2.014 & 1.133 \\
        & & 720 & 0.424 & 0.439 & 0.459 & 0.465 & 0.829 & 0.693 & 0.492 & 0.480 & 0.499 & 0.497 & 2.355 & 1.294 \\
        \cmidrule{3-15}
        & & Avg & 0.379 & 0.405 & 0.398 & 0.422 & \textbf{0.518} & \textbf{0.504} & \textbf{0.501} & \textbf{0.482} & \textbf{0.438} & \textbf{0.445} & \textbf{0.675} & \textbf{1.722} \\
        \midrule
        \end{tabular}
    }
    \caption{Full results of VCA generality experiments on six Transformer-based models.}
    \label{tab:full_generality}
\end{table*}

\end{document}